\documentclass[10pt,journal,compsoc]{IEEEtran}
\usepackage{amsmath,amsfonts}
\usepackage{array}
\usepackage[caption=false,font=normalsize,labelfont=sf,textfont=sf]{subfig}
\usepackage{textcomp}
\usepackage{stfloats}
\usepackage{url}
\usepackage{verbatim}
\usepackage{graphicx}
\usepackage{cite}
\usepackage{balance}
\usepackage{stfloats}

\usepackage{amsmath}
\usepackage{multirow}
\usepackage{booktabs}
\usepackage{longtable}
\usepackage[ruled,vlined]{algorithm2e}
\usepackage[none]{hyphenat}
\usepackage[figuresright]{rotating}

\usepackage{xcolor}

\usepackage{scalerel}
\usepackage{tikz}
\usetikzlibrary{svg.path}

\definecolor{orcidlogocol}{HTML}{A6CE39}
\tikzset{
    orcidlogo/.pic={
        \fill[orcidlogocol] svg{M256,128c0,70.7-57.3,128-128,128C57.3,256,0,198.7,0,128C0,57.3,57.3,0,128,0C198.7,0,256,57.3,256,128z};
        \fill[white] svg{M86.3,186.2H70.9V79.1h15.4v48.4V186.2z}
        svg{M108.9,79.1h41.6c39.6,0,57,28.3,57,53.6c0,27.5-21.5,53.6-56.8,53.6h-41.8V79.1z M124.3,172.4h24.5c34.9,0,42.9-26.5,42.9-39.7c0-21.5-13.7-39.7-43.7-39.7h-23.7V172.4z}
        svg{M88.7,56.8c0,5.5-4.5,10.1-10.1,10.1c-5.6,0-10.1-4.6-10.1-10.1c0-5.6,4.5-10.1,10.1-10.1C84.2,46.7,88.7,51.3,88.7,56.8z};
    }
}

\newcommand\orcidicon[1]{\href{https://orcid.org/#1}{\mbox{\scalerel*{
                \begin{tikzpicture}[yscale=-1,transform shape]
                \pic{orcidlogo};
                \end{tikzpicture}
            }{|}}}}

\usepackage[implicit=false]{hyperref}
\hypersetup{hidelinks,
	colorlinks=true,
	allcolors=black,
	pdfstartview=Fit,
	breaklinks=true}

\begin{document}
%
\title{Reinforced Lin-Kernighan-Helsgaun Algorithms for the Traveling Salesman Problems}
%
%
%
%

\author{Jiongzhi~Zheng,~Kun~He${\textsuperscript{\orcidicon{0000-0001-7627-4604}}}$,~\IEEEmembership{Senior~Member,~IEEE,} Jianrong~Zhou,~Yan~Jin, and~Chu-Min~Li
\IEEEcompsocitemizethanks{\IEEEcompsocthanksitem Jiongzhi Zheng, Kun He, Jianrong Zhou, and Yan Jin are with the School of Computer Science and Technology, Huazhong University of Science and Technology, Wuhan 430074, China. E-mail: \{jzzheng, brooklet60, jrzhou\}@hust.edu.cn, jinyan@mail.hust.edu.cn.
\IEEEcompsocthanksitem Chu-Min Li is with the MIS, University of Picardie Jules Verne, Amiens 80000, France. E-mail: chu-min.li@u-picardie.f.}
\thanks{Corresponding author: Kun He.}}

\IEEEtitleabstractindextext{%
\begin{abstract}
TSP is a classical NP-hard combinatorial optimization problem with many practical variants. LKH is one of the state-of-the-art local search algorithms for the TSP. LKH-3 is a powerful extension of LKH that can solve many TSP variants. Both LKH and LKH-3 associate a candidate set to each city to improve the efficiency, and have two different methods, $\alpha$-measure and POPMUSIC, to decide the candidate sets. In this work, we first propose a Variable Strategy Reinforced LKH (VSR-LKH) algorithm, which incorporates three reinforcement learning methods (Q-learning, Sarsa, Monte Carlo) with LKH, for the TSP. We further propose a new algorithm called VSR-LKH-3 that combines the variable strategy reinforcement learning method with LKH-3 for typical TSP variants, including the TSP with time windows (TSPTW) and Colored TSP (CTSP). The proposed algorithms replace the inflexible traversal operations in LKH and LKH-3 and let the algorithms learn to make a choice at each search step by reinforcement learning. Both LKH and LKH-3, with either $\alpha$-measure or POPMUSIC, can be significantly improved by our methods. Extensive experiments on 236 widely-used TSP benchmarks with up to 85,900 cities demonstrate the excellent performance of VSR-LKH. VSR-LKH-3 also significantly outperforms the state-of-the-art heuristics for TSPTW and CTSP.

\end{abstract}

\begin{IEEEkeywords}
Traveling salesman problems, reinforcement learning, Lin-Kernighan-Helsgaun algorithm, local search
\end{IEEEkeywords}}

\maketitle

\IEEEdisplaynontitleabstractindextext

%
\IEEEpeerreviewmaketitle


\IEEEraisesectionheading{\section{Introduction}\label{Sec-Intro}}
\IEEEPARstart{G}{iven} a set of cities with certain locations, the Traveling Salesman Problem (TSP) is to find the shortest Hamiltonian route, along which a salesman travels from a city (depot) to visit all the cities exactly once and finally returns to the depot. The TSP and its variant problems, such as the TSP with time windows (TSPTW)~\cite{Silva2010,Karabulut2014} and the colored TSP (CTSP)~\cite{Li2015,He2021GMA}, are all challenging combinatorial optimization problems. Algorithms designed for the TSP and its variant problems have many possible applications, such as the Vehicle Routing Problem (VRP)~\cite{Nazari2018}, collision-free scheduling of multi-bridge machining systems~\cite{Li2018}, rice harvesting schedules~\cite{He2018}, etc.

Numerous approaches have been proposed for solving the TSP and its variant problems. Typical methods mainly include exact algorithms, approximation algorithms, and heuristic algorithms. Exact algorithms may be prohibitive for large-scale instances and approximation algorithms may suffer from weak optimality guarantees or empirical performance~\cite{Khalil2017}. Heuristics are known to be the most efficient and effective approaches for solving the TSPs.

Since the Lin-Kernighan (LK) heuristic~\cite{Lin1973} was proposed, the LK-based algorithms~\cite{Lin1973,Martin1991,Helsgaun2000,Taillard2019} have become one of the most popular series of local search heuristics for the TSP. Among them, the Lin-Kernighan-Helsgaun (LKH) algorithm~\cite{Helsgaun2000,Taillard2019} is the most famous and also one of the state-of-the-art local search algorithms for the TSP. LKH optimizes the solutions by the $k$-opt method~\cite{Lin1965}, which explores the solution space by replacing $k$ edges of the current tour. To reduce the search scope and improve the efficiency, LKH defines the candidate edges as LK does and restricts that the edges to be added during the $k$-opt process must be selected from the candidate edges. LKH associates a candidate set for each city, where it records several other candidate cities. The candidate edges in LKH consist of the edges between each city and its candidate cities.

In LK, the candidate set of each city records several other nearest cities. The earliest version of LKH~\cite{Helsgaun2000} proposes an $\alpha$-measure method to determine the candidate sets, which uses an $\alpha$-value defined based on the minimum spanning tree~\cite{Held1970} as a metric for evaluating the quality of the edges. The $\alpha$-measure further improves the performance of the $\alpha$-value by adding a $\pi$-value as penalties to each city. We denote the $\alpha$-value obtained after adding $\pi$-values as $\alpha^{\pi}$-value for clarity. Due to the better performance of the $\alpha^{\pi}$-value than the distance in determining the candidate cities, LKH~\cite{Helsgaun2000} significantly improves the LK heuristic.

Recently, Taillard and Helsgaun~\cite{Taillard2019} propose to apply the Partial OPtimization Metaheuristic Under Special Intensification Conditions (POPMUSIC) method to help LKH select the candidate edges. POPMUSIC is an efficient approach that can quickly generate high-quality solutions. Its main idea is to divide a solution into multiple sub-paths, and improve each sub-paths independently. LKH with the POPMUSIC method~\cite{Taillard2019} first uses POPMUSIC to generate multiple tours, and then determines the candidate edges by merging all the edges that appeared in these tours. This approach can obtain the candidate sets within a shorter time than the $\alpha$-measure method~\cite{Helsgaun2000}, simultaneously guaranteeing the quality of the candidate sets. After determining the candidate sets by POPMUSIC, the order of the candidate cities is sorted according to the $\alpha$-value.

LKH-3~\cite{Helsgaun2017} is an extension of LKH that can be used to solve about 40 different TSP variants. LKH-3 solves these problems by transforming them into the constrained TSP~\cite{Jonker1986,Rao1980,Jonker1988}, and uses the $k$-opt method to explore the solution space. LKH-3 allows searching in the infeasible solution space, and defines different violation functions for different problems to evaluate the violation extent of the given constraints. A solution is improved by $k$-opt in LKH-3 when the violation function is reduced, or the violation function is unchanged meanwhile the optimization objective is reduced. A route with zero violation values is feasible. The candidate sets in LKH-3 can also be selected by both $\alpha$-measure and POPMUSIC.

Both LKH and LKH-3 are successful and effective algorithms. However, the constant $\alpha^{\pi}$-value or $\alpha$-value may not be the best metric to determine or order the candidate cities, although they are well designed. During the $k$-opt process, both LKH and LKH-3 traverse the candidate sets in ascending order of the $\alpha^{\pi}$-value (or $\alpha$-value) to select the edges to be added. Such a traversal mechanism is inflexible and may limit the potential of the algorithms to find better solutions. In this work, we address the challenge of improving LKH and LKH-3, and introduce an effective method to combine reinforcement learning~\cite{Sutton1998,Zou2021,Li2022} with the LKH and LKH-3 algorithms. The proposed algorithms could learn to adjust the candidate sets adaptively and learn to choose appropriate edges to be added during the $k$-opt process by means of reinforcement learning methods.

Specifically, we first propose a Variable Strategy Reinforced LKH (VSR-LKH) algorithm, that combines three reinforcement learning methods, Q-learning, Sarsa, and Monte Carlo~\cite{Sutton1998}, with LKH to solve the TSP. The reinforcement learning is combined with the core $k$-opt process in LKH. VSR-LKH actually learns an adaptive Q-value to replace the $\alpha^{\pi}$-value (or $\alpha$-value) as the metric for evaluating the quality of the edges. Moreover, the variable strategy reinforcement learning method inspired from Variable Neighborhood Search (VNS)~\cite{Mladenovic1997} could make use of the complementary of the above three reinforcement learning methods in reinforcing LKH (for example, some TSP instances can be solved well by Q-learning, but not by Sarsa, and vice versa). We then combine the variable strategy reinforcement learning method with LKH-3 to solve two typical variants of the TSP, TSPTW and CTSP. The resulting algorithm is denoted as VSR-LKH-3.

This paper is an extended and improved version of our conference paper~\cite{Zheng2021}, in which we proposed the VSR-LKH algorithm for the standard TSP and compared with LKH with the $\alpha$-measure method. In this paper, we make a more comprehensive empirical analysis of the VSR-LKH for the TSP, and further propose a new algorithm called VSR-LKH-3 that can handle various TSP variants. Due to the properties of the violation functions in LKH-3, we design a new reward function in the reinforcement learning framework of VSR-LKH-3 that considers both tour lengths and violation values. We do a more comprehensive comparison with LKH and LKH-3 with both the $\alpha$-measure and the POPMUSIC methods. We further compare VSR-LKH with the state-of-the-art (deep) learning based algorithm for the TSP, NeuroLKH~\cite{Xin2021}, and compare VSR-LKH-3 with the state-of-the-art heuristic algorithms for the TSPTW~\cite{Silva2010,Karabulut2014} and CTSP~\cite{He2021GMA}. 

The principal contributions of this work are as follows:
\begin{itemize}
\item We propose to combine reinforcement learning with the effective LKH and LKH-3 algorithms to solve the TSP and its variants. Both LKH and LKH-3, with either the $\alpha$-measure or the POPMUSIC method can be significantly improved by our methods. 
\item VSR-LKH also significantly outperforms the state-of-the-art (deep) learning based algorithm for the TSP, NeuroLKH. Moreover, VSR-LKH-3 significantly outperforms the state-of-the-art heuristics for solving the TSPTW and CTSP. In particular, our method can help LKH-3 outperform the best-performing CTSP heuristic, providing 12 new best solutions among 65 public CTSP instances.
\item We define a Q-value to replace the $\alpha^{\pi}$-value (or $\alpha$-value) used in LKH and LKH-3 for selecting and sorting the candidate cities. The Q-value can be adjusted adaptively by learning from the information of the solutions generated during the iterative search.
\item Since heuristics for NP-hard combinatorial optimization problems usually need to search in the solution space, our approach suggests a method to improve  heuristics by combining reinforcement learning with the core local search process.
\end{itemize}

The rest of this paper is organized as follows. 
Section \ref{Sec-Prob} introduces basic concepts for the TSP, TSPTW, and CTSP. Section \ref{Sec-RW} presents related works of effective heuristics and (reinforcement) learning based methods in solving the TSPs, and further discusses the advantages of our proposed reinforcement learning method. Section \ref{Sec-LKH} briefly introduces the existing LKH and LKH-3 algorithms. Sections \ref{Sec-VSRLKH} and \ref{Sec-VSRLKH3} describes our proposed VSR-LKH and VSR-LKH-3 algorithms. Section \ref{Sec-Exp} presents experimental results and analyses. Section \ref{Sec-Con} contains the concluding remarks.

\section{Problem Definition}
\label{Sec-Prob}
This section presents the definition of the problems to be addressed, TSP, TSPTW, and CTSP, respectively.

\subsection{Traveling Salesman Problem}
Given a complete, undirected graph $G=(V,E)$, where $V = \{1,2,...,n\}$ denotes the set of the cities and $E$ the set of all pairwise edges $\{(i,j)| i,j \in V\}$. Each edge $(i,j) \in E$ has a cost $d(i,j)$, e.g., the distance of traveling from city $i$ to city $j$. The TSP is to find a Hamiltonian route represented by a permutation $(c_1, c_2,..., c_n)$ of cities $\{1, 2, ..., n\}$ that minimizes the total cost, i.e., $d(c_1,c_2)+d(c_2,c_3)+...+d(c_n,c_1)$, where $c_1=1$ is the depot.

\subsection{Traveling Salesman Problem with Time Windows}
In the TSP with Time Windows (TSPTW), each city $i \in {V}$ has a time window $[a_i,b_i]$ that restricts the salesman from reaching city $i$ within the time window, and a service time $st_i$ that indicates the time length the salesman must stay in city $i$. In addition, we assume that waiting is permitted. If the salesman reaches city $i$ before $a_i$, he will wait until time $a_i$ to start the service. The TSPTW is to find the shortest Hamiltonian route on the graph $G = (V, E)$ while satisfying the time window constraints.

\subsection{Colored Traveling Salesman Problem}
\label{Sec-Prob-CTSP}
In the Colored TSP (CTSP), the city set $V$ is divided into $m+1$ disjoint sets: $m$ exclusive city sets $Ec_1,Ec_2,...,Ec_m$ and one shared city set $Sc$. The cities of each exclusive set $Ec_k (k=1,2, ...,m)$ must be visited by salesman $k$ and each city from the shared city set $Sc$ can be visited by any of the $m$ salesmen. City 1 (the depot) belongs to the shared set $Sc$ and is visited by all the salesmen. The CTSP needs to determine $m$ routes on the graph $G = (V, E)$ for the $m$ salesmen. Route $k$ ($k \in \{1,2,...,m\}$) can be represented by sequence ($c_1^k,c_2^k,...c_{l_k}^k$), where $c_1^k=1$ is the depot, $l_k$ is the number of cities in route $k$. The $m$ routes should satisfy the following constraints. First, each city except the depot can be visited exactly once, i.e., $\cup_{k=1}^m{\cup_{i=2}^{l_k}{\{c_i^k\}}}=V \backslash \{1\}$ and $\sum\nolimits_{k=1}^m{(l_k - 1)}=|V \backslash \{1\}|$. Second, the cities belonging to exclusive set $Ec_k$ should be contained in sequence ($c_1^k,c_2^k,...c_{l_k}^k$), i.e., $\cup_{i=1}^{l_k}{c_i^k}\cap{Ec_k}=Ec_k$. The goal of the CTSP is to find $m$ routes with the minimum total cost, i.e., $\sum\nolimits_{k=1}^m{d(c_1^k,c_2^k)+d(c_2^k,c_3^k)+...+d(c_{l_k}^k,c_1^k)}$.

\section{Related Work}
\label{Sec-RW}
This section reviews related works of effective heuristic algorithms and (reinforcement) learning based methods in solving the TSP, TSPTW, and CTSP.

\subsection{Heuristic Algorithms}
The LK series of algorithms~\cite{Lin1973,Martin1991,Helsgaun2000,Taillard2019} introduced in Section~\ref{Sec-Intro} are a kind of state-of-the-art heuristics for the TSP based on local search. Additionally, the edge assembly crossover genetic algorithm (EAX-GA)~\cite{Nagata2006,Nagata2013} is another kind of state-of-the-art TSP heuristics based on the genetic algorithms. EAX-GA is very efficient in solving TSP instances with tens of thousands of cities. However, it is hard to scale to super large instances, such as TSP instances with millions of cities, since the convergence of the population is too time-consuming and the memory required for storing the information of the population is huge. As an efficient local search algorithm, LKH can yield near-optimal solutions faster as compared with EAX-GA. Moreover, LKH is suitable for TSP instances with various scales, especially for super large instances, providing the best-known solution of the famous World TSP instance with 1,904,711 cities\footnote{http://www.math.uwaterloo.ca/tsp/world/index.html}. This paper mainly focuses on improving the LKH and LKH-3 local search algorithms by reinforcement learning.

Effective heuristics for the TSPTW include the compressed-annealing heuristic~\cite{Ohlmann2007}, the general variable neighborhood search (GVNS) algorithm~\cite{Silva2010}, and the variable iterated greedy variable neighborhood search (VIG\_VNS) algorithm~\cite{Karabulut2014}. The compressed-annealing heuristic augments the temperature from traditional simulated annealing with the concept of pressure, and proposes a variable penalty function and stochastic search to allow the algorithm to search in infeasible solution space. The GVNS algorithm consists of a construction phase and an optimization phase. During the optimization phase, GVNS uses the variable neighborhood descend algorithm to improve the initial solution generated in the construction phase. The VIG\_VNS algorithm proposes a destruction and construction procedure based on iterative greedy algorithms to explore the solution space and find better solutions.

The CTSP was first proposed in 2015~\cite{Li2015}, where the authors presented a genetic algorithm with dual-chromosome coding to solve the problem. Subsequently, some population-based heuristics~\cite{Meng2018,Pandiri2018} and a variable neighborhood search method~\cite{Meng2018-VNS} were proposed. More recently, He \textit{et al.} proposed an effective iterated two-phase local search algorithm~\cite{He2021} and a grouping memetic algorithm (GMA)~\cite{He2021GMA} that combines the EAX~\cite{Nagata2013} crossover method with local search approaches. Among these heuristics, the GMA algorithm based on the EAX crossover method significantly outperforms the others.

\subsection{Learning based Methods}
Since there are few (reinforcement) learning based methods for the TSPTW and CTSP as compared to those for the TSP, this subsection mainly introduces the (reinforcement) learning based methods for the TSP.

(Reinforcement) learning based methods for the TSP can be divided into two categories. The first category solves the problems in an end-to-end manner. Algorithms belonging to this category are usually based on deep neural networks. When receiving an input instance, they use the trained learning model to generate a solution directly. For example, Bello \textit{et al.}~\cite{Bello2017} address TSP by using the actor-critic method to train a pointer network. The S2V-DQN algorithm~\cite{Khalil2017} applies reinforcement learning to train a graph neural network so as to solve several combinatorial optimization problems including the TSP. Goh \textit{et al.}~\cite{Goh2022} use an encoder based on a standard multi-headed transformer architecture and a Softmax or Sinkhorn~\cite{Cuturi2013} decoder to directly solve the TSP. There are also some algorithms based on deep reinforcement learning for variants of the TSPTW~\cite{Zhang2020,Wu2021}. These methods provide good innovations in the field of applying machine learning to solve combinatorial optimization problems. They can yield near-optimal or optimal solutions for small TSP instances with less than hundreds of cities. However, they are usually hard to scale to large instances (with more than thousands of cities) due to the complexity of deep neural networks.

Methods belonging to the second category combine (reinforcement) learning methods with traditional algorithms. Some of them use traditional algorithms as the core and frequently call the learning models to make decisions. For example, Liu and Zeng~\cite{Liu2009} employ reinforcement learning to construct mutation individuals in the previous version of EAX-GA~\cite{Nagata2006} and report better results than EAX-GA and LKH on instances with up to 2,392 cities. But the efficiency of their proposed algorithm is not as good as that of LKH. Costa \textit{et al.}~\cite{Costa2020} and Sui \textit{et al.}~\cite{Sui2021} use deep reinforcement learning to guide 2-opt and 3-opt local search operators, and report results on instances with no more than 500 cities. Other methods separate the learning models and traditional algorithms. They first apply learning models to yield initial solutions~\cite{Zhao2021} or some configuration information~\cite{Xin2021}, and then use traditional algorithms to find high-quality solutions followed the obtained initial solutions or information. Among them, the NeuroLKH algorithm~\cite{Xin2021} is one of the state-of-the-art, which uses a Sparse Graph Network with supervised learning to generate the candidate edges for LKH. It reports better or similar results compared with LKH in instances with less than 6,000 cities. However, the supervised learning method makes its performance depend much on the structure of the training instances, the generated fixed candidate edges may limit its flexibility and robustness, and the complexity of the networks requires huge resources for large instances (such as instances with tens of thousands cities).

In summary, (reinforcement) learning based methods with deep neural networks for the TSP may suffer from the bottleneck of hardly solving large scale instances, and the combination of traditional reinforcement learning methods (training tables, not deep neural networks) with existing (heuristic) algorithms may reduce the algorithm efficiency. Our proposed reinforcement learning method can avoid these issues. On the one hand, instead of using deep neural networks, our reinforcement learning method uses the traditional reinforcement algorithms including Q-learning, Sarsa, and Monte Carlo~\cite{Sutton1998} to train a table that stores the adaptive Q-value to guide the search directions during the $k$-opt process. On the other hand, we combine reinforcement learning with the core search steps of LKH and LKH-3 in a reasonable way, that can adjust the candidate sets adaptively. Therefore, our proposed algorithms can solve very large instances efficiently and effectively, such as the TSP instance with 85,900 cities and CTSP instances with 7,397 cities used in our experiments.

\section{Revisiting the LKH and LKH-3 Algorithms}
\label{Sec-LKH}
This section first introduces the LKH algorithm, and then introduces the extended implementation of LKH-3 over LKH. Our proposed VSR-LKH and VSR-LKH-3 algorithms are implemented based on them.

\subsection{The LKH algorithm}
The main framework of LKH is simple. In each iteration, LKH first generates an initial solution, and then uses the $k$-opt method~\cite{Lin1965} to improve the solution to a local optimum. The edges to be added during the $k$-opt process in LKH are restricted by the candidate sets. This subsection introduces the core approaches used in LKH, including the $k$-opt process and the methods for constructing the candidate sets, i.e., $\alpha$-measure and POPMUSIC.

\subsubsection{The $k$-opt Process in LKH}
\

\begin{figure}[t]
\centering
\includegraphics[width=0.4\columnwidth]{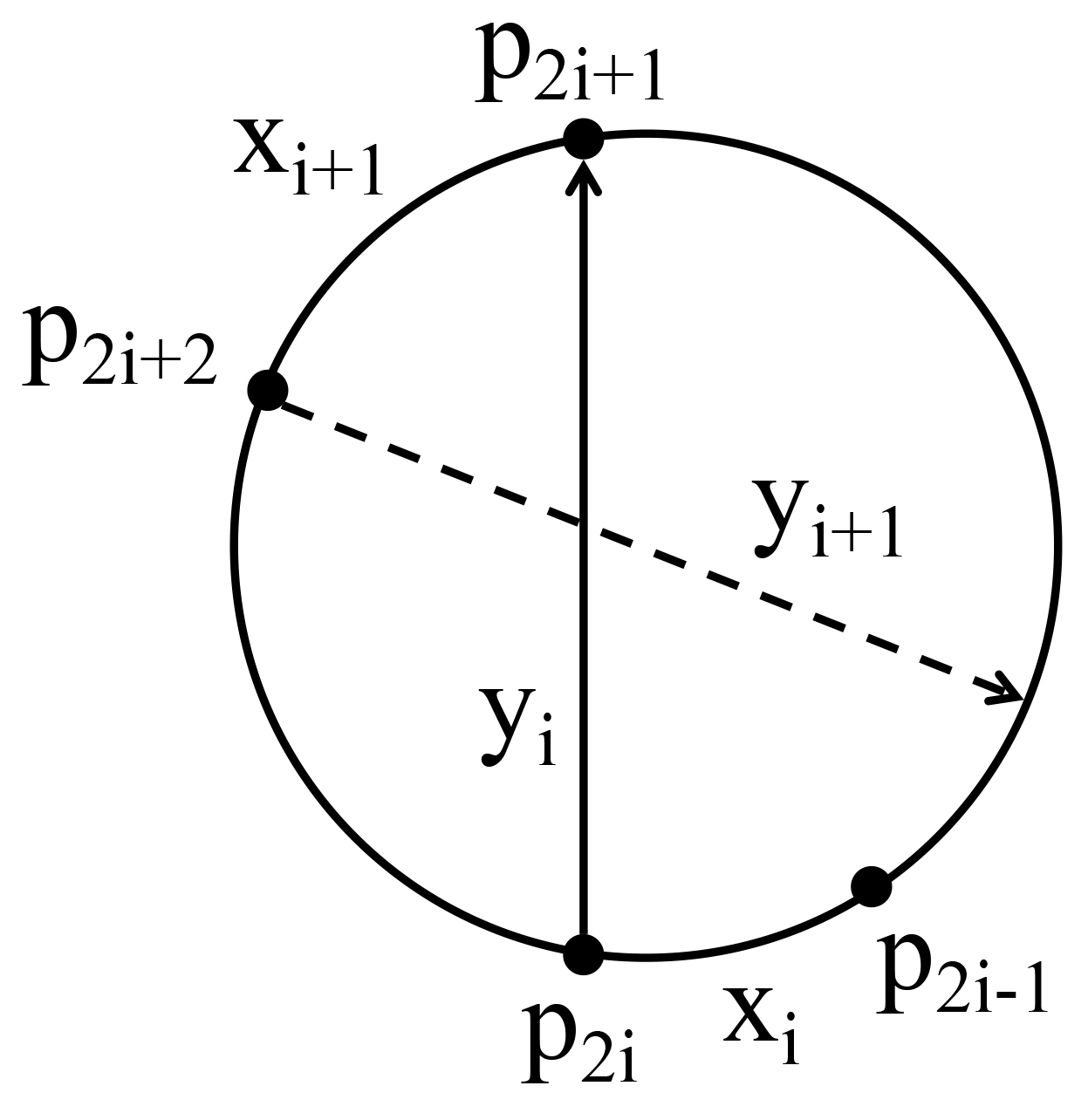}
\vspace{-1em}
\caption{Regarding edges involved in $k$-opt as a sequence of cities.}
\label{fig_xy}
\end{figure}

The $k$-opt method tries to improve the current solution by replacing $k$ edges in the current TSP tour with $k$ new edges. The key point of $k$-opt is how to select the edges to be removed $\{x_1,...,x_k\}$ and the edges to be added $\{y_1,...,y_k\}$. In LKH, the edges involved in a $k$-opt move are connected. Thus selecting the edges can be regarded as selecting a sequence (cycle) of cities, i.e., $(\mathbf{p}_1,\mathbf{p}_2,...,\mathbf{p}_{2k},\mathbf{p}_{2k+1})$ as depicted in Figure \ref{fig_xy}, where $(\mathbf{p}_{2i-1},\mathbf{p}_{2i}) = x_i$, $(\mathbf{p}_{2i},\mathbf{p}_{2i+1}) = y_i$ ($i \in \{1,...,k\}$), and $\mathbf{p}_{2k+1} = \mathbf{p}_1$.

The $k$-opt process in LKH is actually a partial depth-first search process, that the maximum depth of the search tree is restricted to $k_{max}$, i.e., $k$-opt can replace at most $k_{max}$ (5 by default in LKH) edges in the current tour. Specifically, $k$-opt starts from a starting city $\mathbf{p}_1$ (i.e., root of the search tree), then alternatively selects an edge to be removed, i.e., edge $(\mathbf{p}_{2i-1},\mathbf{p}_{2i})$, and an edge to be added, i.e., edge $(\mathbf{p}_{2i},\mathbf{p}_{2i+1})$, until the entire search tree is traversed or a $k$-opt move that can improve the current tour is found. The selection of the cities $\mathbf{p}_{2i}$ and $\mathbf{p}_{2i+1}$ should satisfy the following constraints:

\begin{itemize}
\item \textbf{C-\uppercase\expandafter{\romannumeral1}:} for $i\geq2$, connecting $\mathbf{p}_{2i}$ back to $\mathbf{p}_1$ should result in a feasible TSP tour.
\item \textbf{C-\uppercase\expandafter{\romannumeral2}:} $\mathbf{p}_{2i+1}$ is always chosen so that $\sum\nolimits_{j=1}^{i}(d(\mathbf{p}_{2j-1},\mathbf{p}_{2j})-d(\mathbf{p}_{2j},\mathbf{p}_{2j+1}))>0$.
\end{itemize}

\begin{algorithm}[t]
\caption{$k$-opt($S_{in},\mathbf{p}_1,\mathbf{p},i,k_{max}$)}
\label{alg_kopt}
\LinesNumbered
\KwIn{input solution: $S_{in}$, starting city: $\mathbf{p}_1$, sequence of the involved cities: $\mathbf{p}$, current search depth: $i$, the maximum search depth: $k_{max}$}
\KwOut{output solution $S_{out}$, sequence of the involved cities $\mathbf{p}$}
\For {$h \leftarrow 1 : 2$}{
$\mathbf{p}_{2i} \leftarrow \mathbf{p}_{2i-1}^{h}$\;
\If{$\mathbf{p}_{2i}$ does not satisfy the constraint C-\uppercase\expandafter{\romannumeral1}}{\textbf{continue\;}}
\If{$i \geq 2 \wedge \sum_{j=1}^{i}{d(\mathbf{p}_{2j-1},\mathbf{p}_{2j})} < \sum_{j=1}^{i-1}{d(\mathbf{p}_{2j},\mathbf{p}_{2j+1})} + d(\mathbf{p}_{2i},\mathbf{p}_1)$}{
$S_{out} \leftarrow S_{in}$\;
\For{$j \leftarrow 1 : i$}{remove edge $(\mathbf{p}_{2j-1},\mathbf{p}_{2j})$ from $S_{out}$\;}
\For{$j \leftarrow 1 : i-1$}{add edge $(\mathbf{p}_{2j},\mathbf{p}_{2j+1})$ into $S_{out}$\;}
Add edge $(\mathbf{p}_{2i},\mathbf{p}_1)$ into $S_{out}$\;
\textbf{return} $(S_{out},\mathbf{p})$\;
}
\lIf{$i = k_{max}$}{\textbf{return} $(S_{in},\emptyset)$}
\For {$j \leftarrow 1 : 5$}{
$\mathbf{p}_{2i+1} \leftarrow$ the $j$-th city in $CS^{\mathbf{p}_{2i}}$\;
\If{$\mathbf{p}_{2i+1}$ doesn't satisfy constraint C-\uppercase\expandafter{\romannumeral2}}{\textbf{continue\;}}
$(S_{temp},\mathbf{p'}) \leftarrow k$-opt($S_{in},\mathbf{p}_1,\mathbf{p} \cup \{\mathbf{p}_{2i},\mathbf{p}_{2i+1}\},i+1,k_{max}$)\;
\If{$l(S_{temp}) < l(S_{in})$}{\textbf{return} $(S_{temp},\mathbf{p'})$\;}
}
}
\textbf{return} $(S_{in},\emptyset)$\;
\end{algorithm}

Let $c^1$ be a city randomly picked from the two cities connected with city $c$ in the current TSP tour, $c^2$ be the other, $l(S)$ be the length of solution $S$, $CS^c$ be the candidate set of city $c$. The procedure of the $k$-opt process is presented in Algorithm \ref{alg_kopt}. As shown in Algorithm \ref{alg_kopt}, the $k$-opt process tries to improve the current solution by traversing the partial depth-first search tree from the root $\mathbf{p}_1$. When selecting the edge to be removed, i.e., edge $(\mathbf{p}_{2i-1},\mathbf{p}_{2i})$ (the same as selecting $\mathbf{p}_{2i}$ from $\mathbf{p}_{2i-1}$), the algorithm traverses the two cities connected with city $\mathbf{p}_{2i-1}$ in the current TSP tour (lines 1-2). When selecting the edge to be added, i.e., edge $(\mathbf{p}_{2i},\mathbf{p}_{2i+1})$ (the same as selecting $\mathbf{p}_{2i+1}$ from $\mathbf{p}_{2i}$), the algorithm traverses the candidate set of city $\mathbf{p}_{2i}$ (lines 14-15). The constraint C-\uppercase\expandafter{\romannumeral2} is applied as a smart pruning strategy to improve the efficiency (lines 16-17). Once a $k$-opt move that can improve the current solution is found, the algorithm performs this move on $S_{in}$ and outputs the resulting solution $S_{out}$ (lines 5-12).

\subsubsection{The $\alpha$-measure}
\

Obviously, the performance of the $k$-opt process in LKH highly depends on the quality of the candidate sets. With the help of the $\alpha$-measure method, the LKH algorithm~\cite{Helsgaun2000} significantly outperforms the previous LK-based algorithms~\cite{Lin1973,Martin1991}.

The $\alpha$-measure method first proposes an $\alpha$-value to evaluate the quality of the edges. The $\alpha$-value is defined from the structure of 1-tree~\cite{Held1970}. A 1-tree for the graph $G = (V, E)$ is a spanning tree on the node set $V\backslash\{v\}$ combined with two edges from $E$ incident to a node $v$ chosen arbitrarily. The minimum 1-tree is the 1-tree with the minimum length. Obviously, the length of the minimum 1-tree is a lower bound of the optimal TSP solution. The $\alpha$-value of an edge $(i,j)$ can be calculated by:
\begin{equation}
\alpha(i,j) = L(T^+(i,j))-L(T),
\label{eq_alpha-value}
\end{equation}
where $L(T)$ is the length of the minimum 1-tree of graph $G$, and $L(T^{+}(i,j))$ is the length of the minimum 1-tree required to contain edge $(i,j)$. 

To further enhance the quality of the $\alpha$-value, LKH uses a method to maximize the lower bound of the optimal TSP solution by adding penalties ($\pi$-values)~\cite{Held1971}. Specifically, a $\pi$-value is added to each city as penalties when calculating the distance between two cities. The $\pi$-values actually change the cost matrix of the TSP. Note that this change does not change the optimal solution of the TSP, but it changes the minimum 1-tree. Suppose $L(T_\pi)$ is the length of the minimum 1-tree after adding the $\pi$-values, then the lower bound $w(\pi)$ of the optimal solution can be calculated by Eq. \ref{eq_w-pi}, which is a function of set $\pi = [\pi_1, ..., \pi_n]$:
\begin{equation}
w(\pi)=L(T_\pi)-2\sum\limits_i\pi_i.
\label{eq_w-pi}
\end{equation}

The $\pi$-values are computed by maximizing the lower bound $w(\pi)$ of the optimal solution using a sub-gradient optimization method~\cite{Held1970}. After adding the $\pi$-values, the $\alpha$-value is further improved for determining the candidate set. We denote the $\alpha$-value obtained after adding the $\pi$-values as $\alpha^{\pi}$-value for distinguish. Finally, the candidate set of each city generated by the $\alpha$-measure method records five (default value) other cities with the smallest $\alpha^{\pi}$-value to this city in ascending order.

\subsubsection{POPMUSIC in LKH}
\

POPMUSIC is an efficient method for solving large TSP instances~\cite{Taillard2019}. Its basic idea is to divide an initial TSP tour into several sub-paths and then improve each sub-path independently. When improving a sub-path, the first and last cities are fixed, by which the sub-paths can be easily merged to generate a Hamiltonian route. As a result, the POPMUSIC can generate near-optimal solutions efficiently for large TSP instances.

Taillard \textit{et al.}~\cite{Taillard2019} propose a method based on POPMUSIC for selecting the candidate cities. That is, calculating multiple tours by the POPMUSIC method, and then collecting all the edges appeared in these tours to generate the candidate sets. Specifically, suppose city $c$ is linked with cities $c_1,...,c_i$ in the collected edges, then the candidate set of city $c$ contains cities $c_1,...,c_{i'}$, $i' = max(5,i)$. The cities in each candidate set are also sorted according to the $\alpha$-value (without adding the $\pi$-values). Note that when using the $\alpha$-measure to generate the candidate sets, the most running time is used for maximizing the value of $w(\pi)$ (Eq. \ref{eq_w-pi}), which is time-consuming for solving large TSP instances. Therefore, using the POPMUSIC method can reduce the calculation time of generating the candidate sets without reducing the quality of the candidate sets.

\subsection{The LKH-3 Algorithm}
The LKH-3 algorithm~\cite{Helsgaun2017} is an extension of LKH for variants of the TSP. It uses the transformation methods~\cite{Jonker1986,Rao1980,Jonker1988} to transform constrained asymmetric TSP and constrained multiple TSP (MTSP) with a single depot into constrained symmetric TSP, and proposes a violation function for each problem to allow the algorithm to search in the infeasible solution space. Note that LKH-3 can also use the $\alpha$-measure or the POPMUSIC method to generate the candidate sets as LKH does. This subsection introduces the main differences in LKH-3.

\subsubsection{Transformations}
\

An asymmetric problem with $n$ cities could be transformed into a symmetric problem with $2n-1$ cities using the transformation method of Jonker and Volgenant~\cite{Jonker1986}. The problems with time windows are first transformed by LKH-3 into asymmetric problems before they are transformed into symmetric problems. 

The vehicle routing problems and some constrained TSPs such as CTSP are variants of MTSP. A symmetric TSP with $m$ salesmen can be solved using a symmetric TSP augmented with $m - 1$ copies of the depot, where infinite costs are assigned between depots~\cite{Rao1980,Jonker1988}. LKH-3 uses the transformation methods proposed by Rao~\cite{Rao1980} as well as Jonker and Volgenant~\cite{Jonker1988} to transform an MTSP with a single depot into a standard TSP.

\subsubsection{Violation Functions}
\

The LKH-3 uses violation functions to handle the constraints and allow the algorithm to search in the infeasible solution space. A violation function returns how much the route violates the given constraints. A route with zero violation values is a feasible solution. In LKH-3, the violation value is separated from the objective value. Associated with each solution is a pair $(f^v, f^o)$, where $f^v$ is the violation value of the solution, and $f^o$ is its objective value (e.g., tour length). Let $S_1$ and $S_2$ be two solutions with associated pairs $(f^v_1, f^o_1)$ and $(f^v_2, f^o_2)$. Then $S_1$ is better than $S_2$ if $(f^v_1<f^v_2) \vee ({f^v_1=f^v_2} \wedge{f^o_1<f^o_2})$. In other words, minimizing the violation value is the primary objective of LKH-3, whereas minimizing cost is secondary. A violation function has been implemented for each problem 
variant in LKH-3. For example, the violation function in TSPTW returns how much the route violates the time window constraints, and the violation function in CTSP returns how much the route violates the exclusive city sets constraints (described in Section \ref{Sec-Prob-CTSP}).

\section{The VSR-LKH Algorithm for the TSP}
\label{Sec-VSRLKH}
We propose a Variable Strategy Reinforced LKH (VSR-LKH) algorithm that combines reinforcement learning with LKH to solve the TSP. In VSR-LKH, we propose a Q-value trained by reinforcement learning to replace the constant $\alpha^{\pi}$-value (or $\alpha$-value) used in LKH. Thus our algorithm can learn to select appropriate edges to be added from the candidate sets. Moreover, we use a variable strategy method that combines the advantages of three reinforcement learning methods (Q-learning, Sarsa, Monte Carlo) to further improve the flexibility and the robustness of the algorithm.

In the following, we first introduce our reinforcement learning framework, then introduce the design of the initial Q-value and the generation of the initial candidate sets in VSR-LKH, finally introduce the reinforcement learning algorithms in VSR-LKH and the main process of VSR-LKH.

\subsection{Reinforcement Learning Framework in VSR-LKH}
Since reinforcement learning in VSR-LKH aims to help the algorithm learn to make correct decisions for choosing the edges to be added during the $k$-opt process, the states and actions in the reinforcement learning framework are all related to the edges to be added, and an episode corresponds to a $k$-opt process, which can be represented by $(S',\mathbf{p}) \leftarrow k$-opt($S,\mathbf{p}_1,\{\mathbf{p}_1\},1,k_{max}$) (see Algorithm \ref{alg_kopt}). The detailed description of the states, actions, and rewards in the reinforcement learning framework are as follows.

\begin{itemize}
\item \textbf{States}: The current state of the system is a city that is going to select an edge to be added. For example, cities $\mathbf{p}_{2i} (i \in \{1,2,...,\frac{|\mathbf{p}|}{2}-1\})$ in sequence $\mathbf{p}$. 
\item \textbf{Actions}: For a state $\mathbf{p}_{2i}$, the action is the selection of another endpoint of the edge to be added except $\mathbf{p}_{2i}$ from the candidate set of $\mathbf{p}_{2i}$. For example, cities $\mathbf{p}_{2i+1} (i \in \{1,2,...,\frac{|\mathbf{p}|}{2}-1\})$ in sequence $\mathbf{p}$. 
\item \textbf{Transition}: The next state after performing an action in the current state is the next city that needs to select an edge to be added. For example, $\mathbf{p}_{2i+2}$ is the state transferred to after taking action $\mathbf{p}_{2i+1}$ at state $\mathbf{p}_{2i}$. 
\item \textbf{Rewards}: The reward function should be able to represent the improvement of the tour when taking an action at the current state. The reward $r(\mathbf{p}_{2i},\mathbf{p}_{2i+1})$ obtained by performing action $\mathbf{p}_{2i+1}$ at state $\mathbf{p}_{2i}$ can be calculated by:
\begin{equation}
r(\mathbf{p}_{2i},\mathbf{p}_{2i+1}) = d(\mathbf{p}_{2i-1},\mathbf{p}_{2i}) - d(\mathbf{p}_{2i},\mathbf{p}_{2i+1}),
\label{eq_r1}
\end{equation}
since the $k$-opt move replaces edge $(\mathbf{p}_{2i-1},\mathbf{p}_{2i})$ with edge $(\mathbf{p}_{2i},\mathbf{p}_{2i+1})$.
\end{itemize}

From the framework we can observe that, a state-action pair $(\mathbf{p}_{2i},\mathbf{p}_{2i+1})$ in sequence $\mathbf{p}$ is actually an edge between city $\mathbf{p}_{2i}$ and its candidate city $\mathbf{p}_{2i+1}$. The Q-value in VSR-LKH is actually an estimated value of the state-action value function, and is also an evaluation metric of an edge like $\alpha^{\pi}$-value, $\alpha$-value, and distance. VSR-LKH learns the Q-value by reinforcement learning and uses the Q-value to replace the $\alpha^{\pi}$-value or $\alpha$-value used in LKH.

\subsection{Initial Estimation of State-action Value Function}
The Q-value in VSR-LKH needs a reasonable initial value. The definitions of the initial Q-value for the $\alpha$-measure and POPMUSIC methods are different. For an edge $(i,j)$, the initial Q-value $Q(i,j)$ for the $\alpha$-measure and POPMUSIC methods is deﬁned respectively as follows:
\begin{equation}
Q(i,j)=\frac{w(\pi)}{\alpha^{\pi}(i,j)+d(i,j)},
\label{eq_QA}
\end{equation}
\begin{equation}
Q(i,j)=\frac{L(T)}{\alpha(i,j)+d(i,j)},
\label{eq_QP}
\end{equation}
where $w(\pi)$ (see Eq. \ref{eq_w-pi}) and $L(T)$ (see Eq. \ref{eq_alpha-value}) are the lower bounds of the optimal TSP solution with and without $\pi$-values, respectively.

The initial Q-value defined in Eq. \ref{eq_QA} and Eq. \ref{eq_QP} combines the factors of selecting and sorting the candidate cities in LK and LKH, i.e., the distance and $\alpha^{\pi}$-value (or $\alpha$-value). Note that $\alpha^{\pi}$-value (or $\alpha$-value) is based on a minimum 1-tree that is rather a global property, while the distance between two cities is a local property. Combining them can take advantage of both properties. Although the influence of the distance factor is small, it can avoid the denominator of being zero. The purpose of $w(\pi)$ or $L(T)$ is two-fold. First, it can prevent the initial Q-value from being much smaller than the rewards. Second, it can adaptively adjust the initial Q-value for different instances.

\subsection{Generating Initial Candidate Sets}
\label{sec_initCS}
LKH with the $\alpha$-measure method selects and sorts the candidate cities according to the $\alpha^{\pi}$-value. LKH with the POPMUSIC method selects the candidate cities by POPMUSIC and sorts the candidate cities according to the $\alpha$-value. VSR-LKH with the $\alpha$-measure method selects and sorts the candidate cities according to the initial Q-value calculated by Eq. \ref{eq_QA}. VSR-LKH with the POPMUSIC method selects the candidate cities by POPMUSIC and sorts the candidate cities according to the initial Q-value calculated by Eq. \ref{eq_QP}. Note that LKH sorts the candidate cities in ascending order of the $\alpha^{\pi}$-value or $\alpha$-value. VSR-LKH sorts the candidate cities in descending order of the Q-value. When initializing the candidate sets, VSR-LKH (resp. LKH) with the $\alpha$-measure method needs to calculate the initial Q-value (resp. $\alpha^{\pi}$-value) of each edge $(i,j) \in E$, and VSR-LKH (resp. LKH) with the POPMUSIC method only needs to calculate the initial Q-value (resp. $\alpha$-value) of the candidate edges selected by POPMUSIC.

Moreover, the experimental results show that the performance of LKH with either the $\alpha$-measure or the POPMUSIC method can be improved significantly by only replacing their initial candidate sets with ours that were generated by our initial Q-value, indicating that the designed initial Q-value is reasonable and effective.

\begin{algorithm}[t]
\caption{VSR-LKH($I_{max},T_{max},k_{max},N_{max},\lambda_1,\gamma_1$)}
\label{alg_VSRLKH}
\LinesNumbered
\KwIn{the maximum number of iterations: $I_{max}$, the cut-off time: $T_{max}$, the maximum search depth of $k$-opt: $k_{max}$, variable strategy parameters: $N_{max}$, learning rate: $\lambda_1$, reward discount factor: $\gamma_1$}
\KwOut{output solution: $S_{best}$}
\textit{Create\_Candidate\_Set}()\;
Initialize the length of the best solution $l(S_{best}) \leftarrow +\infty$\, $Trial \leftarrow 0$, $num \leftarrow 0$\;
Initialize $M \leftarrow 1$ that corresponds to the algorithms (1: Q-learning, 2: Sarsa, 3: Monte Carlo)\;
\While{$Trial < I_{max} \wedge$ running time $< T_{max}$}{
$Trial  \leftarrow Trial + 1$, $num  \leftarrow num + 1$\;
\If{$num\geq{N_{max}}$}{$M \leftarrow M\%{3}+1,num \leftarrow 0$\;}
$S_{init} \leftarrow$ \textit{Choose\_Initial\_Tour}(), $S_{curr} \leftarrow S_{init}$\;
Initialize the set of cities that have not been selected as the starting city of the $k$-opt: $A \leftarrow \{1,2,...,n\}$\;
\While{TRUE}{
\lIf{$A = \emptyset$}{\textbf{break}}
$\mathbf{p}_1 \leftarrow$ a random city in $A$, $A \leftarrow A \backslash\{\mathbf{p}_1\}$\;
$(S_{temp},\mathbf{p}) \leftarrow k$-opt($S_{curr},\mathbf{p}_1,\{\mathbf{p}_1\},1,k_{max}$)\;
Update the Q-value of each state-action pair in $\mathbf{p}$ by Eq. \ref{eq_MC}, \ref{eq_Sarsa}, or \ref{eq_QLearning} according to $M$\;
\If{$l(S_{temp}) < l(S_{curr})$}{
$S_{curr} \leftarrow S_{temp}$\;
\lFor{$j \leftarrow 1 : |\mathbf{p}|$}{$A \leftarrow A \cup \{\mathbf{p}_j\}$}
}
}
Sort the candidate sets of each city in descending order of the Q-value\;
\If{$l(S_{curr}) < l(S_{best})$}{$S_{best} \leftarrow S_{curr}$, $num \leftarrow 0$\;}
}
\textbf{return} $S_{best}$\;
\end{algorithm}

\subsection{Reinforcement Learning Algorithms in VSR-LKH}
VSR-LKH applies reinforcement learning to learn to adjust the Q-value so as to estimate the state-action value function more accurately. We choose the widely-used Monte Carlo method and one-step Temporal-Difference (TD) algorithms including Q-learning and Sarsa~\cite{Sutton1998} to train the Q-value during the $k$-opt process. 

\begin{itemize}
\item \textbf{Monte Carlo}:
Monte Carlo is a well-known model-free reinforcement learning method that estimates the value function by averaging the sampled returns. In the reinforcement learning framework, there are no repetitive state-action pairs in one episode, i.e., a $k$-opt move. Therefore, for any state action pair $(\mathbf{p}_{2i},\mathbf{p}_{2i+1})$ in sequence $\mathbf{p}$, the Monte Carlo method uses the episode return after taking action $\mathbf{p}_{2i+1}$ at state $\mathbf{p}_{2i}$ as the estimation of its Q-value. That is, 
\begin{equation}
Q(\mathbf{p}_{2i},\mathbf{p}_{2i+1})=\sum\nolimits_{t=0}^{\frac{|\mathbf{p}|}{2}-1}{r(\mathbf{p}_{2(i+t)},\mathbf{p}_{2(i+t)+1})},
\label{eq_MC}
\end{equation}
\item \textbf{One-step TD}:
TD learning is a combination of Monte Carlo and dynamic programming. The TD algorithms can update the Q-values in an online, fully incremental fashion. In this work, we use both the on-policy TD control (Sarsa) and off-policy TD control (Q-learning) to reinforce the LKH. The one-step Sarsa and Q-learning update the Q-value respectively as follows:
\begin{equation}
\begin{aligned}
Q(\mathbf{p}_{2i},&\mathbf{p}_{2i+1})=(1-\lambda_1)\cdot{Q(\mathbf{p}_{2i},\mathbf{p}_{2i+1}}+\\
\lambda_1&\cdot[r(\mathbf{p}_{2i},\mathbf{p}_{2i+1})+\gamma_1{Q(\mathbf{p}_{2i+2},\mathbf{p}_{2i+3})}],
\label{eq_Sarsa}
\end{aligned}
\end{equation}
\begin{equation}
\begin{aligned}
Q(\mathbf{p}_{2i},&\mathbf{p}_{2i+1})=(1-\lambda_1)\cdot{Q(\mathbf{p}_{2i},\mathbf{p}_{2i+1}}+\\
\lambda_1&\cdot[r(\mathbf{p}_{2i},\mathbf{p}_{2i+1})+\gamma_1{\max\limits_{a' \in CS^{\mathbf{p}_{2i+2}}}Q(\mathbf{p}_{2i+2},a')}],
\label{eq_QLearning}
\end{aligned}
\end{equation}
where $\lambda_1$ and $\gamma_1$ are the learning rate and the reward discount factor in VSR-LKH, respectively.
\end{itemize}

\subsection{Main Process of VSR-LKH}
Now we introduce the main process of VSR-LKH, as presented in Algorithm \ref{alg_VSRLKH}. The \textit{Create\_Candidate\_Set}() function (line 1) creates the initial candidate sets according to the method described in Section \ref{sec_initCS}. At each iteration (lines 4-20), an initial solution $S_{init}$ is generated (line 8) by the initialization method in LKH (i.e., function \textit{Choose\_Initial\_Tour}()), which generates the initial solution by perturbing the best solution found so far, i.e., $S_{best}$. VSR-LKH uses the $k$-opt heuristic (Algorithm \ref{alg_kopt}) to improve the current solution $S_{curr}$ until reaching the local optimum (lines 10-17), i.e., $S_{curr}$ cannot be improved by the $k$-opt heuristic starting from any starting city $\mathbf{p}_1 \in \{1,2,...,n\}$.

During the $k$-opt process in VSR-LKH, the reinforcement learning algorithms are used to update the Q-value (line 14), so as to adjust the candidate sets to make them better (line 18). A variable strategy method is applied to combine the advantages of the three reinforcement learning algorithms, Q-learning, Sarsa, and Monte Carlo, and leverage their complementarity. Specifically, when the best solution $S_{best}$ has not been improved for $N_{max}$ iterations with the current reinforcement learning algorithm, the variable strategy mechanism will switch to another algorithm (lines 6-7). VSR-LKH stops when the maximum number of iterations or the cut-off time is reached (line 4).

Note that after generating the initial candidate sets, VSR-LKH only needs to store the Q-value of each candidate edge $(i,j) \in \{(i,j) \in E | j \in CS^i \vee i \in CS^j\}$ as LKH does, which stores the $\alpha^{\pi}$-value or $\alpha$-value of each candidate edge. Thus, VSR-LKH only needs a table with the size of around $5 \times n$ to store the Q-values.

\section{VSR-LKH-3 for TSP Variants}
\label{Sec-VSRLKH3}
We further propose a VSR-LKH-3 algorithm that combines the variable strategy reinforcement learning method with LKH-3 to solve typical variants of the TSP. VSR-LKH-3 also applies the Q-value adapted by reinforcement learning to replace the $\alpha^{\pi}$-value or $\alpha$-value in LKH-3. The reinforcement learning framework, the initial Q-value, and the reinforcement learning algorithms used in VSR-LKH-3 are the same as those in VSR-LKH. The differences between VSR-LKH-3 and VSR-LKH are mainly on the reward function. Intuitively, since LKH-3 considers the violation function as the primary optimization objective, the reward function in VSR-LKH-3 should consider both the tour length and the violation function. Experimental results also demonstrate that considering the violation function in the reward function is effective and reasonable.

Note that VSR-LKH-3 can solve 38 different variants of the TSP as LKH-3 does. Among these variants, there are about 8 versions with time window constraints and 21 versions with multiple salesmen. Therefore, we select two typical variants, TSPTW and CTSP, one with time window constraints and another with multiple salesmen, to empirically evaluate the proposed VSR-LKH-3 in our experiments. The results show that VSR-LKH-3 significantly outperforms LKH-3, with either the $\alpha$-measure or the POPMUSIC method, in solving the TSPTW and CTSP. Thus intuitively, VSR-LKH-3 can also yield better results than LKH-3 in solving the other variants of the TSP with time window constraints or multiple salesmen.

The rest of this section introduces the reward function, the reinforcement learning algorithms, and the main process of VSR-LKH-3, respectively.

\subsection{Reward Function in VSR-LKH-3}
Suppose $(\mathbf{p}_{2i},\mathbf{p}_{2i+1})$ is a state-action pair in sequence $\mathbf{p}$ (i.e., an episode), $r(\mathbf{p}_{2i},\mathbf{p}_{2i+1})$ is the reward function in VSR-LKH (see Eq. \ref{eq_r1}) that indicates the influence of performing action $\mathbf{p}_{2i+1}$ and state $\mathbf{p}_{2i}$ on the tour length, $\Delta f^v(\mathbf{p}_{2i},\mathbf{p}_{2i+1})$ is the reduction of the violation value of a route after performing action $\mathbf{p}_{2i+1}$ and state $\mathbf{p}_{2i}$. Since the violation value is the extent of the constraint violation of a solution, it cannot be calculated during the $k$-opt process. Thus the values of $\Delta f^v(\mathbf{p}_{2i},\mathbf{p}_{2i+1})$ for each state-action pair $(\mathbf{p}_{2i},\mathbf{p}_{2i+1})$, $i \in \{1,...,\frac{\mathbf{p}}{2}-1\}$, in an episode are the same and equal to the violation reduction of the route after the overall $k$-opt process, denoted simply as $\Delta f^v$. The reward function in VSR-LKH-3 should consider both $\Delta f^v$ and $r(\cdot,\cdot)$, since both the violation function and the tour length are optimization objectives in LKH-3 when solving the constrained TSPs.

To distinguish with reward function in VSR-LKH, we denote $r'(\mathbf{p}_{2i},\mathbf{p}_{2i+1})$ as the reward obtained by performing action $\mathbf{p}_{2i+1}$ and state $\mathbf{p}_{2i}$.
Intuitively,  $r'(\mathbf{p}_{2i},\mathbf{p}_{2i+1})$ should be positive when $\Delta f^v>0$ or $\Delta f^v=0\wedge{r(\mathbf{p}_{2i},\mathbf{p}_{2i+1})>0}$, and be negative when $\Delta f^v<0$ or $\Delta f^v=0\wedge{r(\mathbf{p}_{2i},\mathbf{p}_{2i+1})<0}$, since the violation value has a higher priority than the objective value in LKH-3. Thus whether the reward is positive or negative is mainly determined by the violation reduction $\Delta f^v$. However, we cannot simply set $r'(\mathbf{p}_{2i},\mathbf{p}_{2i+1})=\Delta f^v$ because using $\Delta f^v$ alone cannot reflect the impact of different values of $r(\mathbf{p}_{2i},\mathbf{p}_{2i+1})$ on rewards, nor can it handle the situation when $\Delta f^v=0$. Therefore, we apply $r(\mathbf{p}_{2i},\mathbf{p}_{2i+1})$ as an additional item for the reward $r'(\mathbf{p}_{2i},\mathbf{p}_{2i+1})$ when $\Delta f^v=0$ to decide the positive or negative as well as the extent of the reward, and when $\Delta f^v>0\wedge{r(\mathbf{p}_{2i},\mathbf{p}_{2i+1})>0}$ or $\Delta f^v<0\wedge{r(\mathbf{p}_{2i},\mathbf{p}_{2i+1})<0}$ as an increment in rewards or penalties. Thus the reward $r'(\mathbf{p}_{2i},\mathbf{p}_{2i+1})$ after taking action $\mathbf{p}_{2i}$ at state $\mathbf{p}_{2i+1}$ in VSR-LKH-3 can be calculated by:
\begin{equation}
\begin{aligned}
r'(&\mathbf{p}_{2i},\mathbf{p}_{2i+1})=\\
&\begin{cases}\Delta f^v+r(\mathbf{p}_{2i},\mathbf{p}_{2i+1})&\text{$\Delta f^v\cdot{r(\mathbf{p}_{2i},\mathbf{p}_{2i+1})}\geq 0$}\\
\Delta f^v&\text{$\Delta f^v\cdot{r(\mathbf{p}_{2i},\mathbf{p}_{2i+1})}<0$}\end{cases},
\label{eq_r2}
\end{aligned}
\end{equation}
where $r(\mathbf{p}_{2i},\mathbf{p}_{2i+1})$ can be calculated by Eq. \ref{eq_r1}.

\subsection{Reinforcement Learning Algorithms in VSR-LKH-3}
VSR-LKH-3 also applies Monte Carlo, Sarsa, and Q-learning algorithms to update the Q-value. They update the Q-value of each state-action pair $(\mathbf{p}_{2i},\mathbf{p}_{2i+1})$, $i \in \{1,...,\frac{\mathbf{p}}{2}-1\}$, in an episode respectively as follows:
\begin{equation}
Q(\mathbf{p}_{2i},\mathbf{p}_{2i+1})=\sum\nolimits_{t=0}^{\frac{|\mathbf{p}|}{2}-1}{r'(\mathbf{p}_{2(i+t)},\mathbf{p}_{2(i+t)+1})},
\label{eq_MC2}
\end{equation}
\begin{equation}
\begin{aligned}
Q(\mathbf{p}_{2i},&\mathbf{p}_{2i+1})=(1-\lambda_2)\cdot{Q(\mathbf{p}_{2i},\mathbf{p}_{2i+1}}+\\
\lambda_2&\cdot[r'(\mathbf{p}_{2i},\mathbf{p}_{2i+1})+\gamma_2{Q(\mathbf{p}_{2i+2},\mathbf{p}_{2i+3})}],
\label{eq_Sarsa2}
\end{aligned}
\end{equation}
\begin{equation}
\begin{aligned}
Q(\mathbf{p}_{2i},&\mathbf{p}_{2i+1})=(1-\lambda_2)\cdot{Q(\mathbf{p}_{2i},\mathbf{p}_{2i+1}}+\\
\lambda_2&\cdot[r'(\mathbf{p}_{2i},\mathbf{p}_{2i+1})+\gamma_2{\max\limits_{a' \in CS^{\mathbf{p}_{2i+2}}}Q(\mathbf{p}_{2i+2},a')}],
\label{eq_QLearning2}
\end{aligned}
\end{equation}
where $\lambda_2$ and $\gamma_2$ in Eq. \ref{eq_Sarsa2} and Eq. \ref{eq_QLearning2} are the learning rate and the reward discount factor in VSR-LKH-3, respectively.

\subsection{Main Process of VSR-LKH-3}
The main flow of VSR-LKH-3 is similar to that of VSR-LKH (Algorithm \ref{alg_VSRLKH}). The differences in VSR-LKH-3 are as follows. First, the equations for updating the Q-value in VSR-LKH-3 are Eqs. \ref{eq_MC2}, \ref{eq_Sarsa2}, and \ref{eq_QLearning2}. Second, the $k$-opt in VSR-LKH-3 terminates the partial depth-first search when finding a $k$-opt move that can reduce the violation function of the current solution or reduce the tour length of the current solution without increasing its violation function.

\section{Experimental Results}
\label{Sec-Exp}
Experimental results provide insight on why and how the proposed approaches are effective, suggesting that our designed initial Q-value that combines the distance and $\alpha^{\pi}$-value (or $\alpha$-value) can improve the performance of LKH and LKH-3. Moreover, the algorithms can be further improved through the variable strategy reinforced learning method. The source codes of VSR-LKH and VSR-LKH-3 are available at https://github.com/JHL-HUST/VSR-LKH-V2.

VSR-LKH and VSR-LKH-3 are implemented in C++ and compiled by g++ with -O3 option. The experiments were run on a server using an Intel® Xeon® E5-2650 v3 2.30 GHz 10-core CPU and 256 GB RAM, running Ubuntu 16.04 Linux operation system. We denote the LKH (VSR-LKH) algorithm with the $\alpha$-measure method as LKH-$\alpha$ (VSR-LKH-$\alpha$), and the LKH (VSR-LKH) algorithm with the POPMUSIC method as LKH-P (VSR-LKH-P). The LKH and LKH-3 baselines in the experiments are their newest versions.

The rest of this section presents the experimental results and analyses for the TSP and its typical variants, including TSPTW and CTSP, respectively.

\subsection{Experimental Results and Analyses for the TSP}
In this subsection, we first introduce the experimental setup including the baseline algorithms, the benchmark instances we used, and the settings of the parameters, and then present the computational results of VSR-LKH and the baseline algorithms. In the end, we analyze the effectiveness of the components in VSR-LKH, including the initial Q-value, the reinforcement learning method, and the variable reinforcement learning strategy.

\subsubsection{Experimental Setup for the TSP}
\

The baseline algorithms for the TSP include the state-of-the-art local search algorithm, LKH, and the state-of-the-art (deep) learning based algorithm, NeuroLKH~\cite{Xin2021}. NeuroLKH uses a Sparse Graph Network with supervised learning to generate the candidate edges for LKH. Thus its main framework is also based on LKH. We compare VSR-LKH with two versions of NeuroLKH. The first is NeuroLKH\_R, which is trained on instances with uniformly distributed nodes. The second is NeuroLKH\_M, which is trained on a mixture of instances with uniformly distributed nodes, clustered nodes, half uniform and half clustered nodes.

We test our VSR-LKH algorithm on all the instances with no more than 85,900 cities, with a total of 236, in the well-known and widely used benchmark sets for the TSP: TSPLIB\footnote{http://comBKSifi.uni-heidelberg.de/software/TSPLIB95}, National TSP benchmarks\footnote{http://www.math.uwaterloo.ca/tsp/world/countries.html}, and VLSI TSP benchmarks\footnote{http://www.math.uwaterloo.ca/tsp/vlsi/index.html}. Note that the number in an instance name indicates the number of cities in that instance. The number of cities of the tested 236 TSP instances ranges from 14 to 85,900. Each instance is solved 10 times by each algorithm as LKH~\cite{Helsgaun2000} does.

In order to make a clear comparison, We divide the 236 tested instances into \textit{small} and \textit{large} according to their scales. An instance is \textit{small} if it has less than 10,000 cities, and otherwise \textit{large}. There are 193 \textit{small} instances and 43 \textit{large} instances among all the 236 tested instances. Moreover, we further divide the 236 instances into \textit{easy}, \textit{medium}, and \textit{hard} according to their difficulty. An instance is \textit{easy} if LKH-$\alpha$, LKH-P, VSR-LKH-$\alpha$, and VSR-LKH-P can always yield the best-known solution in each of the 10 runs. An instance is \textit{hard} if at least one of LKH-$\alpha$, LKH-P, VSR-LKH-$\alpha$, and VSR-LKH-P can not yield the best-known solution in 10 runs. The rest instances are \textit{medium}. There are 99 \textit{easy} instances, 77 \textit{medium} instances, and 60 \textit{hard} instances among all the 236 tested instances.

We set parameters related to the stopping criterion including the cut-off time $T_{max}$ and the maximum number of iterations $I_{max}$ to be the same for VSR-LKH, LKH, and NeuroLKH. $T_{max}$ is set to $n$ seconds, $I_{max}$ is set to $n$ (the default settings in LKH) for \textit{small} instances and $n/5$ for \textit{large} instances. Parameters related to reinforcement learning in our VSR-LKH algorithm include the learning rate $\lambda_1$, the reward discount factor $\gamma_1$, and the variable strategy parameter $MaxNum$. Their default settings are as follows: $\lambda_1 = 0.1$, $\gamma_1 = 0.9$, and $MaxNum = I_{max} / 20$.

\begin{table*}[thp]
\centering
\caption{Comparison on VSR-LKH-$\alpha$ and VSR-LKH-P with NeuroLKH\_R and NeuroLKH\_M. Best results appear in bold.\vspace{-1em}}
\label{table_NeuroLKH}
\scalebox{0.59}{
\begin{tabular}{lrrrrrrrrrrrrrrrrrrrrr} \bottomrule
\multicolumn{1}{c}{\multirow{2}{*}{Instances}} &  \multirow{2}{*}{\hspace{-1em}BKS} & \hspace{-1em}  &  \multicolumn{4}{c}{\hspace{-1em}NeuroLKH\_R}                               & \hspace{-1em}  &  \multicolumn{4}{c}{\hspace{-1em}NeuroLKH\_M}                                 & \hspace{-1em}  &  \multicolumn{4}{c}{\hspace{-1em}VSR-LKH-$\alpha$}                                   & \hspace{-1em}  &  \multicolumn{4}{c}{\hspace{-1em}VSR-LKH-P}                                   \\ \cline{4-7} \cline{9-12} \cline{14-17} \cline{19-22} 
\multicolumn{1}{c}{}                           & \hspace{-1em}                      & \hspace{-1em}  & \hspace{-1em} Best (\%)                   & \hspace{-1em} Average (\%)           & \hspace{-1em} Time    & \hspace{-1em} Trial  & \hspace{-1em}  & \hspace{-1em} Best  (\%)                      & \hspace{-1em} Average (\%)            & \hspace{-1em} Time    & \hspace{-1em} Trial  & \hspace{-1em}  & \hspace{-1em} Best (\%)                      & \hspace{-1em} Average (\%)            & \hspace{-1em} Time    & \hspace{-1em} Trials & \hspace{-1em}  & \hspace{-1em} Best (\%)                      & \hspace{-1em} Average (\%)            & \hspace{-1em} Time    & \hspace{-1em} Trials \\ \hline
d198                                           & \hspace{-1em} 15780                & \hspace{-1em}  & \hspace{-1em} 15789 (0.0570)          & \hspace{-1em} 15825.0 (0.2852)   & \hspace{-1em} 3.69    & \hspace{-1em} 198.0  & \hspace{-1em}  & \hspace{-1em} \textbf{15780 (0.0000)}   & \hspace{-1em} 15780.0 (0.0000)   & \hspace{-1em} 2.18    & \hspace{-1em} 1.0    & \hspace{-1em}  & \hspace{-1em} \textbf{15780 (0.0000)}   & \hspace{-1em} 15780.0 (0.0000)   & \hspace{-1em} 0.89    & \hspace{-1em} 1.4    & \hspace{-1em}  & \hspace{-1em} \textbf{15780 (0.0000)}   & \hspace{-1em} 15780.0 (0.0000)   & \hspace{-1em} 0.52    & \hspace{-1em} 1.0    \\
\end{tabular}}
\vspace{-1em}
\end{table*}
\begin{table*}[t]
\centering
\caption{Comparison of VSR-LKH-$\alpha$ and LKH-$\alpha$, VSR-LKH-P and LKH-P, on 60 \textit{hard} instances. Best results appear in bold.\vspace{-1em}}
\label{table_TSP}
\scalebox{0.53}{\begin{tabular}{lrrrrrrrrrrr|rrrrrrrrrr} \bottomrule
\multirow{2}{*}{Name} &  \multirow{2}{*}{\hspace{-1em}BKS} &  \multicolumn{4}{c}{\hspace{-1em}LKH-$\alpha$}                                            & \hspace{-1em}  &  \multicolumn{4}{c}{\hspace{-1em}VSR-LKH-$\alpha$}                                        & \hspace{-1em}  & \hspace{-1em} \multirow{63}{*}{} &  \multicolumn{4}{c}{\hspace{-1em}LKH-P}                                            & \hspace{-1em}  &  \multicolumn{4}{c}{\hspace{-1em}VSR-LKH-P}                                       \\ \cline{3-6} \cline{8-11} \cline{14-17} \cline{19-22} 
                      & \hspace{-1em}                      & \hspace{-1em} Best (\%)                  & \hspace{-1em} Average (\%)         & \hspace{-1em} Time     & \hspace{-1em} Trials  & \hspace{-1em}  & \hspace{-1em} Best (\%)                   & \hspace{-1em} Average (\%)         & \hspace{-1em} Time     & \hspace{-1em} Trials & \hspace{-1em}  & \hspace{-1em}                    & \hspace{-1em} Best (\%)                   & \hspace{-1em} Average (\%)         & \hspace{-1em} Time     & \hspace{-1em} Trials & \hspace{-1em}  & \hspace{-1em} Best (\%)                  & \hspace{-1em} Average (\%)         & \hspace{-1em} Time     & \hspace{-1em} Trials \\ \hline
pcb1173               & \hspace{-1em} 56892                & \hspace{-1em} \textbf{56892 (0.0000)}    & \hspace{-1em} 56895.0 (0.0053)     & \hspace{-1em} 2.38     & \hspace{-1em} 152.4   & \hspace{-1em}  & \hspace{-1em} \textbf{56892 (0.0000)}     & \hspace{-1em} 56892.0 (0.0000)     & \hspace{-1em} 9.17     & \hspace{-1em} 319.3  & \hspace{-1em}  & \hspace{-1em}                    & \hspace{-1em} 56897 (0.0088)              & \hspace{-1em} 56897.0 (0.0088)     & \hspace{-1em} 0.54     & \hspace{-1em} 44.8   & \hspace{-1em}  & \hspace{-1em} \textbf{56892 (0.0000)}    & \hspace{-1em} 56892.0 (0.0000)     & \hspace{-1em} 8.15     & \hspace{-1em} 258.5  \\
\end{tabular}}
\vspace{-1em}
\end{table*}
\begin{table}[thp]
\centering
\caption{Summarized comparisons for the TSP.\vspace{-1em}}
\label{table_Compare-TSP}
\scalebox{0.9}{\begin{tabular}{lrrrr} \bottomrule
Algorithm pair                    & Indicator & \#Wins & \#Ties & \#Losses \\ \hline
VSR-LKH-$\alpha$ vs. NeuroLKH\_R  & Best      & 38     & 121     & 1        \\
                                  & Average   & 75     & 75      & 10       \\ \hline
VSR-LKH-$\alpha$ vs. NeuroLKH\_M  & Best      & 20     & 137     & 3        \\
                                  & Average   & 50     & 94     & 16        \\ \hline
VSR-LKH-P vs. NeuroLKH\_R         & Best      & 40     & 120     & 0        \\
                                  & Average   & 80     & 72      & 8         \\ \hline
VSR-LKH-P vs. NeuroLKH\_M         & Best      & 22     & 138     & 0        \\
                                  & Average   & 51     & 93     & 16        \\ \hline
VSR-LKH-$\alpha$ vs. LKH-$\alpha$ & Best      & 43     & 85     & 9        \\
                                  & Average   & 116    & 10     & 11        \\ \hline
VSR-LKH-P vs. LKH-P               & Best      & 24     & 98     & 15       \\
                                  & Average   & 87     & 17     & 33      \\ \toprule
\end{tabular}}
\vspace{-1em}
\end{table}

\begin{sidewaystable}[thp]
\centering
\caption{Comparison of VSR-LKH-3, LKH-3, GVNS, and VIG\_VNS on \textit{Datasets A}, \textit{B}, and \textit{C}. Best results appear in bold. The values in brackets beside some results are the average violation values of these solutions.\vspace{-0.3em}}
\label{table_TSPTW}
\scalebox{0.55}{\begin{tabular}{llrrrrrrrrrrrrrrrrrrrrrrrrr} \bottomrule
\multicolumn{2}{l}{Instances} & \multirow{2}{*}{BKS} &  & \multicolumn{3}{c}{GVNS}            &  & \multicolumn{3}{c}{VIG\_VNS}       &  & \multicolumn{3}{c}{LKH-3-$\alpha$}           &  & \multicolumn{3}{c}{LKH-3-P}                    &  & \multicolumn{3}{c}{VSR-LKH-3-$\alpha$}    &  & \multicolumn{3}{c}{VSR-LKH-3-P}                \\ \cline{1-2} \cline{5-7} \cline{9-11} \cline{13-15} \cline{17-19} \cline{21-23} \cline{25-27} 
$n$                      & $w$    &                      &  & Best             & Average  & Time  &  & Best             & Average  & Time &  & Best             & Average            & Time &  & Best             & Average              & Time &  & Best              & Average        & Time &  & Best             & Average              & Time \\ \hline
\textit{Dataset A}                & -     & 579.50               &  & \textbf{579.50}  & 579.51   & -     &  & \textbf{579.50}  & 579.50   & -    &  & \textbf{579.50}  & 579.50             & -    &  & \textbf{579.50}  & (0.04) 579.33         & -    &  & \textbf{579.50}   & 579.50         & -    &  & \textbf{579.50}  & (0.04) 579.33         & -    \\ \hline
\textit{Dataset B}                & -     & 486.17               &  & 486.23           & 486.48   & -     &  & \textbf{486.17}  & 486.42   & -    &  & \textbf{486.17}  & 486.27             & -    &  & \textbf{486.17}  & 486.24               & -    &  & \textbf{486.17}   & 486.23         & -    &  & \textbf{486.17}  & 486.21               & -    \\ \hline
\textit{Dataset C}              &      &                      &  &                  &          &       &  &                  &          &      &  &                  &                    &      &  &                  &                      &      &  &                   &                &      &  &                  &                      &      \\
\end{tabular}}
\end{sidewaystable}
\begin{table}[thp]
\centering
\caption{Summarized comparisons for the TSPTW.\vspace{-1em}}
\label{table_Compare-TSPTW}
\scalebox{0.95}{\begin{tabular}{lrrrr} \bottomrule
Algorithm pair                        & Indicator & \#Wins & \#Ties & \#Losses \\ \hline
VSR-LKH-3-$\alpha$ vs. GVNS           & Best      & 16     & 68     & 1        \\
                                      & Average   & 36     & 43     & 6        \\ \hline
VSR-LKH-3-$\alpha$ vs. VIG\_VNS       & Best      & 6      & 79     & 0        \\
                                      & Average   & 25     & 52     & 8        \\ \hline
VSR-LKH-3-P vs. GVNS                  & Best      & 15     & 65     & 5        \\
                                      & Average   & 24     & 39     & 22       \\ \hline
VSR-LKH-3-P vs. VIG\_VNS              & Best      & 5      & 74     & 6        \\
                                      & Average   & 14     & 47     & 24       \\ \hline
VSR-LKH-3-$\alpha$ vs. LKH-3-$\alpha$ & Best      & 6      & 78     & 1        \\
                                      & Average   & 27     & 54     & 4        \\ \hline
VSR-LKH-3-P vs. LKH-3-P               & Best      & 7      & 73     & 5        \\
                                      & Average   & 24     & 52     & 9             \\ \toprule 
\end{tabular}}
\vspace{-1em}
\end{table}

\subsubsection{Comparison on VSR-LKH with the Baselines}
\

We first compare VSR-LKH-$\alpha$ and VSR-LKH-P with NeuroLKH\_R and NeuroLKH\_M. Note that the resources required by NeuroLKH are numerous for large scale instances. The performance of NeuroLKH for large instances is also limited due to the small scale of the supervised training instances. Therefore, we only compare VSR-LKH with NeuroLKH on \textit{small} instances with less than 10,000 cities (Xin \textit{et al.} only report results on instances with less than 6,000 cities in~\cite{Xin2021}) and two-dimensional Euclidean distance (EUC\_2D) metric (NeuroLKH only supports the EUC\_2D metric), with a total of 160 instances. We extract the instances that the best solutions of  VSR-LKH-$\alpha$, VSR-LKH-P, NeuroLKH\_R, and NeuroLKH\_M obtained in 10 runs all equal to the best-known solution. The results of the remaining 42 instances are shown in Table \ref{table_NeuroLKH}.

We then compare VSR-LKH-$\alpha$ with LKH-$\alpha$, and compare VSR-LKH-P with LKH-P on all the 236 tested instances. Detailed comparison results on 60 \textit{hard} instances of the algorithms with the $\alpha$-measure method and the POPMUSIC method are shown in Table \ref{table_TSP}. In Tables \ref{table_NeuroLKH} and \ref{table_TSP}, we compare the best and average solutions in 10 runs obtained by the algorithms. Column \textit{BKS} indicates the best-known solution of the corresponding instance, column \textit{Time} is the average calculation time (in seconds) of the algorithms, and column \textit{Trial} is the average number of iterations of the algorithms. The values in the brackets beside the results equal to the gap of the results to the best-known solutions multiplied by 100. We also provide the average gap of the best and average solutions to the best-known solutions.

Moreover, in order to show the advantage of VSR-LKH over the baselines in solving the TSP more clearly, we summarize the comparative results between VSR-LKH and each baseline algorithm in Table \ref{table_Compare-TSP}. Columns \textit{\#Wins}, \textit{\#Ties}, and \textit{\#Losses} indicate the number of instances for which VSR-LKH-$\alpha$ or VSR-LKH-P obtains a better, equal, and worse result than the compared algorithm according to the best solution and average solution indicators. The comparative results between VSR-LKH and NeuroLKH are based on all the 160 compared instances. The comparative results between VSR-LKH and LKH are based on all the 137 \textit{medium} and \textit{hard} instances.

From the results in Tables \ref{table_NeuroLKH}, \ref{table_TSP}, and \ref{table_Compare-TSP} we observe that:

(1) VSR-LKH significantly outperforms NeuroLKH\_R and NeuroLKH\_M. The average gaps of the best solutions and average solutions of VSR-LKH-$\alpha$ and VSR-LKH-P are much smaller than those of NeuroLKH\_R and NeuroLKH\_M. The performance of NeuroLKH\_M is better than that of NeuroLKH\_R, indicating that the performance of NeuroLKH relies on the structure of the training instances. Generating reasonable training instances that help NeuroLKH work well on instances with diverse structures is challenging. Moreover, both NeuroLKH\_R and NeuroLKH\_M are not good at solving large instances, indicating that the bottlenecks in large scale instances still limit algorithms based on deep neural networks.

(2) VSR-LKH significantly outperforms LKH on the best and average solutions in 10 runs when solving most of the 137 \textit{medium} and \textit{hard} instances. Specifically, the best (average) solutions of VSR-LKH-$\alpha$ are better than those of LKH-$\alpha$ in 43 (116) instances, and worse than those of LKH-$\alpha$ in 9 (11) instances. The best (average) solutions of VSR-LKH-P are better than those of LKH-P in 24 (87) instances, and worse than those of LKH-P in 15 (33) instances. The results indicate that under the same settings of $I_{max}$ and $T_{max}$, our proposed VSR-LKH algorithm can yield significantly better results than LKH, with either the $\alpha$-measure or the POPMUSIC method.

\begin{sidewaysfigure}[thp]
\centering
\subfloat[]{\includegraphics[width=0.48\columnwidth]{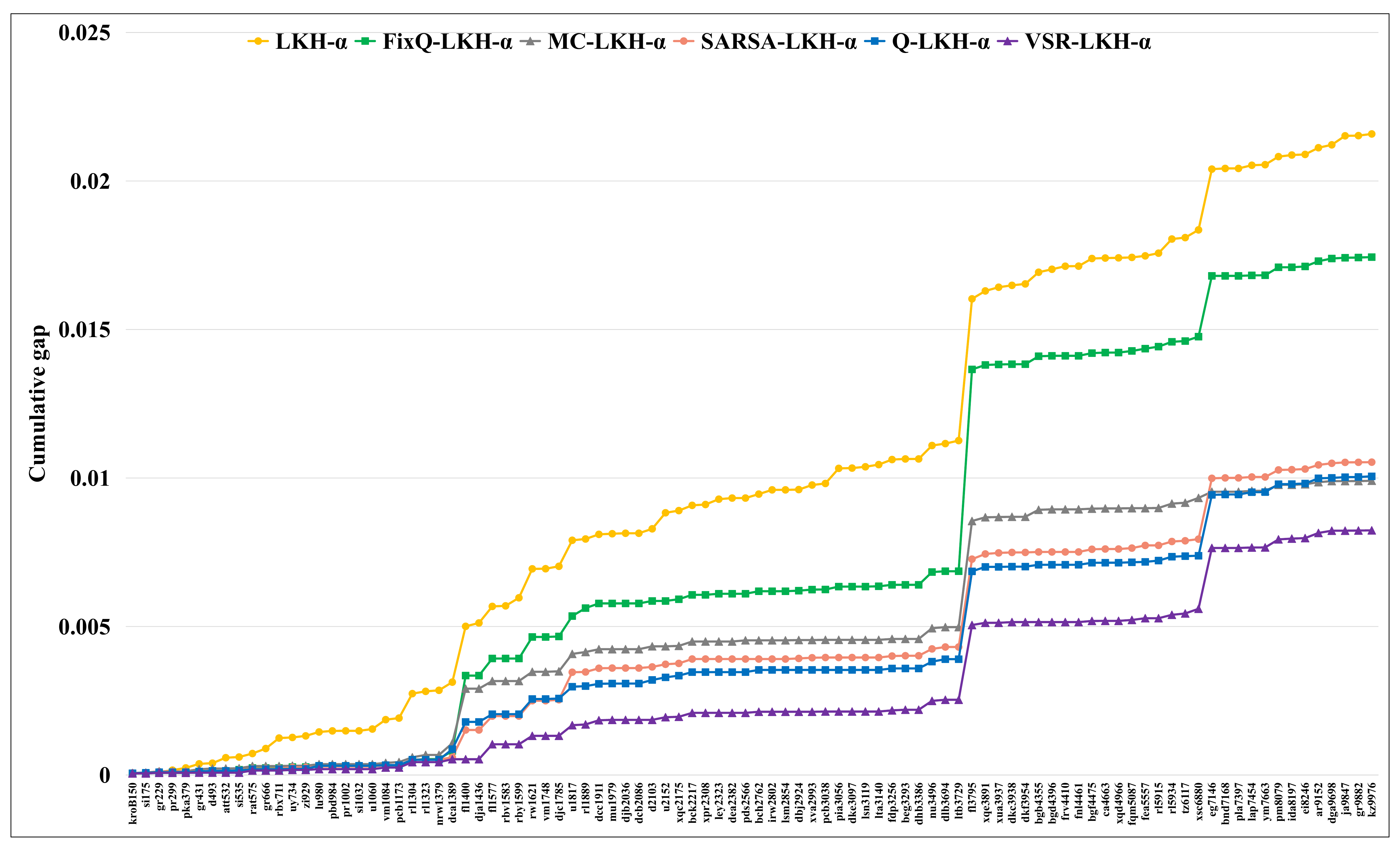}%
\label{fig_VSRLKH-A}}
\hfil
\subfloat[]{\includegraphics[width=0.48\columnwidth]{figs/VSRLKH3-2-A.pdf}%
\label{fig_VSRLKH-P}}
\caption{Comparison on the 94 small and hard TSP instances. Instances are sorted from left to right in ascending order of
the number of the cities. (a) Results of algorithms with the $\alpha$-measure method. (b) Results of algorithms with the POPMUSIC method.}
\label{fig_VSRLKH}
\end{sidewaysfigure}

\subsubsection{Ablation Study for VSR-LKH}
\

In order to evaluate the effectiveness of the components in VSR-LKH, including the initial Q-value, the reinforcement learning method, and the variable reinforcement learning strategy, we compare VSR-LKH and LKH with their several variant algorithms. FixQ-LKH is a variant of LKH that replaces the initial candidate sets of LKH with the initial candidate sets generated according to the initial Q-value (see Section \ref{sec_initCS} for details). Q-LKH, SARSA-LKH, and MC-LKH are three variants of VSR-LKH that only use Q-learning, Sarsa, or Monte Carlo to update the Q-value, i.e., $M$ is fixed to be 1, 2, or 3 in Algorithm \ref{alg_VSRLKH}. We denote FixQ-LKH-$\alpha$ as the FixQ-LKH algorithm with the $\alpha$-measure method and Q-LKH-P as the Q-LKH algorithm with the POPMUSIC method, and so on.

We compare VSR-LKH, Q-LKH, SARSA-LKH, MC-LKH, FixQ-LKH, and LKH for solving the \textit{small} and not \textit{easy} instances, with a total of 94 instances. Figures \ref{fig_VSRLKH-A} and \ref{fig_VSRLKH-P} show the results of the algorithms with $\alpha$-measure and POPMUSIC, respectively. The results are expressed by the cumulative gap on the solution quality. We use the cumulants to express the results because it is more intuitive than comparing the results of each instance individually. Let $gap(j)=\frac{1}{10}\sum_{i=1}^{10}\frac{A_i-BKS_j}{BKS_j}$ be the average gap of calculating the $j$-th instance by an algorithm in 10 runs, where $A_i$ is the result of the $i$-th calculation and $BKS_j$ is the best-known solution of the $j$-th instance. The smaller the average gap, the closer the average solution is to the best-known solution. For an algorithm, $C_{gap}(j)=\sum_{i=1}^{j}gap(i)$ is the cumulative gap.

From the results in Figure \ref{fig_VSRLKH}, we can observe that:

(1) FixQ-LKH shows significantly better performance than LKH, indicating that the LKH algorithm with either the $\alpha$-measure or the POPMUSIC method can be improved significantly by simply replacing the $\alpha^{\pi}$-value or $\alpha$-value with our designed initial Q-value.

(2) Each of Q-LKH, SARSA-LKH, MC-LKH shows better performance than FixQ-LKH, indicating that the initial Q-value can be further improved by our reinforcement learning method with any of the three tested reinforcement learning algorithms. In other words, using an adaptive value trained by reinforcement learning to adjust the candidate sets is better than determining the candidate sets according to a fixed metric.

(3) Q-learning, Sarsa, and Monte Carlo show complementary performance in reinforcing LKH. When solving most of the 94 instances, Q-learning has the best performance in reinforcing LKH, followed by Sarsa and Monte Carlo. Thus we order Q-learning first and Monte Carlo last in VSR-LKH. Moreover, VSR-LKH outperforms Q-LKH, SARSA-LKH, and MC-LKH, indicating that our approach can combine the advantages of the three reinforcement learning methods and leverage their complementarity.

\begin{sidewaystable*}[thp]
\centering
\caption{Comparison of VSR-LKH-3, LKH-3, and GMA on \textit{Datasets \uppercase\expandafter{\romannumeral1}}, \textit{\uppercase\expandafter{\romannumeral2}}, and \textit{\uppercase\expandafter{\romannumeral3}}. Best results appear in bold. The values in brackets beside some results are the average violation values of these solutions. A star in column \textit{Best} indicates that a new best solution of the CTSP has been found.}
\label{table_CTSP}
\scalebox{0.54}{
\begin{tabular}{lrrrrrrrrrrrrrrrrrrrrrrr} \bottomrule
\multirow{2}{*}{Instances} & \multirow{2}{*}{$n$} & \multirow{2}{*}{$m$} & \multirow{2}{*}{BKS} &  & \multicolumn{3}{c}{GMA}                     &  & \multicolumn{3}{c}{LKH-3-$\alpha$}              &  & \multicolumn{3}{c}{LKH-3-P}                     &  & \multicolumn{3}{c}{VSR-LKH-3-$\alpha$}            &  & \multicolumn{3}{c}{VSR-LKH-3-P}                     \\ \cline{6-8} \cline{10-12} \cline{14-16} \cline{18-20} \cline{22-24} 
                      &                    &                    &                      &  & Best               & Average     & Time     &  & Best             & Average           & Time     &  & Best             & Average           & Time     &  & Best               & Average           & Time     &  & Best                 & Average           & Time     \\ \hline
\textit{Dataset \uppercase\expandafter{\romannumeral1}}             &                    &                    &                      &  &                    &             &          &  &                  &                   &          &  &                  &                   &          &  &                    &                   &          &  &                      &                   &          \\ \hline
eil21-2               & 21                 & 2                  & 144.92               &  & \textbf{144.92}    & 144.92      & 0.05     &  & \textbf{144.92}  & 144.92            & 0.00     &  & \textbf{144.92}  & 144.92            & 0.00     &  & \textbf{144.92}    & 144.92            & 0.00     &  & \textbf{144.92}      & 144.92            & 0.00     \\
eil21-3               & 21                 & 3                  & 157.48               &  & \textbf{157.48}    & 157.48      & 0.06     &  & \textbf{157.48}  & 157.48            & 0.00     &  & \textbf{157.48}  & 157.48            & 0.00     &  & \textbf{157.48}    & 157.48            & 0.00     &  & \textbf{157.48}      & 157.48            & 0.00     \\
\end{tabular}}
\end{sidewaystable*}

\subsection{Experimental Results for the TSPTW}
This subsection first introduces the experimental setup, then presents the computational results of VSR-LKH-3 and the baseline algorithms for the TSPTW.

\subsubsection{Experimental Setup}
\

\label{sec_TSPTW_setup}

We compare VSR-LKH-3 with LKH-3 and other state-of-the-art heuristics for the TSPTW, including the GVNS algorithm~\cite{Silva2010} and the VIG\_VNS algorithm~\cite{Karabulut2014}. The results of GVNS and VIG\_VNS are from the literature. The GVNS was run on a Pentium 4 2.40 GHz processor. The VIG\_VNS was run on an Intel Core i5 2.53 GHz processor.

We compare the algorithms in solving the TSPTW on the benchmark instances proposed by~\cite{Ohlmann2007,Silva2010,Dumas1995,Gendreau1998}, with a total of 425 instances. Each instance is solved 10 times by each algorithm. We divide these instances into three sets:

\begin{itemize}
\item \textit{Dataset A:} This dataset~\cite{Dumas1995} contains 135 instances with 20 to 200 cities grouped in 27 test cases. Each group has five Euclidean instances, coordinates between 0 and 50, with the same number of cities and the same maximum range of time windows.
\item \textit{Dataset B:} This dataset~\cite{Ohlmann2007,Gendreau1998} contains 165 instances with 20 to 200 cities grouped in 33 test cases. This set of instances, in the majority, are the instances proposed by Dumas \textit{et al.}~\cite{Dumas1995} with wider time windows, systematically extended by 100.
\item \textit{Dataset C:} This dataset~\cite{Silva2010} contains 125 instances with 200 to 400 cities grouped in 25 test cases. It contains relatively large-scale instances that can distinguish the performance of different heuristics.
\end{itemize}

The cut-off time $T_{max}$ and the maximum number of iterations $I_{max}$ are set to $n/10$ seconds and $+\infty$, respectively, in VSR-LKH-3 and LKH-3 for the TSPTW. The parameters related to reinforcement learning in VSR-LKH-3 include the learning rate $\lambda_2$, the reward discount factor $\gamma_2$, and the variable strategy parameter $N_{max}$. We set $\lambda_2 = 0.1$ and $\gamma_2 = 0.9$ as VSR-LKH does. For the variable strategy parameter, since VSR-LKH-3 does not restrict the maximum number of iterations, we use the running time to control the switching of the reinforcement learning algorithm in VSR-LKH-3. That is, changing the reinforcement learning algorithm when the best solution has not been improved for $T_{max}/20$ seconds.

\subsubsection{Comparison on VSR-LKH-3 with TSPTW Baselines} 
\

The comparison results on VSR-LKH-3 with LKH-3, GVNS, and VIG\_VNS are shown in Table \ref{table_TSPTW}. Since the instances in \textit{Datasets A} and \textit{B} are too simple to distinguish the performance of the algorithms, we only present the average values of the best and average solutions of the algorithms on these two benchmarks in Table \ref{table_TSPTW}. For the instances in \textit{Datasets C}, we present the average values of the best and average solutions of the five instances in each group of instances. The values in brackets beside some results are the average violation values of these solutions. Column $w$ is the width of the time windows. Column \textit{BKS} is the best-known solution reported by~\cite{Silva2010,Karabulut2014,Helsgaun2017}.

In order to show the advantage of VSR-LKH-3 over the baselines in solving the TSPTW more clearly, we summarize the comparative results between VSR-LKH-3 and each baseline algorithm in Table \ref{table_Compare-TSPTW}. The comparative results are based on all the 85 compared groups of instances.

From the results in Tables \ref{table_TSPTW} and \ref{table_Compare-TSPTW}, we can observe that:

(1) The algorithms with the POPMUSIC method show significantly lower performance than the algorithms with the $\alpha$-measure. This might be because the POPMUSIC is not good at solving the problems with time window constraints. Moreover, the results of VSR-LKH-3-P are significantly better than those of LKH-3-P, indicating that our proposed reinforcement learning method and the reward function that considers the penalty function are effective.

(2) When solving the TSPTW, VSR-LKH-3 also significantly outperforms LKH-3 with either the $\alpha$-measure or the POPMUSIC method. Moreover, the best (average) solutions of VSR-LKH-3-$\alpha$ are better than those of GVNS in 16 (36) instances, and worse than those of GVNS in one instance (6 instances). The best (average) solutions of VSR-LKH-3-$\alpha$ are better than those of VIG\_VNS in 6 (25) instances, and worse than those of VIG\_VNS in 0 (8) instances, indicating a significant improvement.

\subsection{Experimental Results for the CTSP}
This subsection first introduces the experimental setup, and then presents the computation results of VSR-LKH-3 and the baseline algorithms for the CTSP.

\begin{table}[t]
\centering
\caption{Summarized comparisons for the CTSP.\vspace{-1em}}
\label{table_Compare-CTSP}
\scalebox{0.95}{\begin{tabular}{lrrrr} \bottomrule
Algorithm pair                        & Indicator & \#Wins & \#Ties & \#Losses \\ \hline
VSR-LKH-3-$\alpha$ vs. GMA            & Best      & 27     & 27     & 11       \\
                                      & Average   & 29     & 20     & 16       \\ \hline
VSR-LKH-3-P vs. GMA                   & Best      & 30     & 27     & 8        \\
                                      & Average   & 28     & 20     & 17       \\ \hline
VSR-LKH-3-$\alpha$ vs. LKH-3-$\alpha$ & Best      & 23     & 38     & 4        \\
                                      & Average   & 29     & 25     & 11       \\ \hline
VSR-LKH-3-P vs. LKH-3-P               & Best      & 21     & 37     & 7        \\
                                      & Average   & 31     & 24     & 10            \\ \toprule 
\end{tabular}}
\vspace{-1em}
\end{table}

\subsubsection{Experimental Setup}
\

There are several effective heuristic algorithms for the CTSP proposed recently~\cite{He2021,Pandiri2018,He2021GMA}. Among these algorithms, the GMA algorithm~\cite{He2021GMA} based on the famous EAX crossover method~\cite{Nagata2013} significantly outperforms the other heuristics. Therefore, we choose LKH-3 and GMA as the baseline algorithms for the CTSP. The results of GMA are obtained by running its source code on our machine.

We tested the algorithms for the CTSP on the benchmark instances introduced by~\cite{Li2015,Pandiri2018,Dong2018,Dong2019} and used by~\cite{He2021,He2021GMA}, with a total of 65 instances. 
Each instance is solved 20 times by each algorithm as in \cite{He2021GMA}. Also, we divide these instances into three sets as in \cite{He2021GMA}:

\begin{itemize}
\item \textit{Dataset \uppercase\expandafter{\romannumeral1}:} This dataset~\cite{Li2015} contains 20 small instances with 21 to 101 cities, generated from six graphs by varying the number of routes and exclusive cities in each instance. The number of salesmen is between 2 and 7.
\item \textit{Dataset \uppercase\expandafter{\romannumeral2}:} This dataset contains 14 medium instances with 202 to 666 cities, generated from four graphs. The number of salesmen is between 10 and 40. The 6 instances related to the two graphs with 202 and 431 cities were from~\cite{Dong2018}, while the remaining instances were presented in~\cite{Pandiri2018}.
\item \textit{Dataset \uppercase\expandafter{\romannumeral3}:} This dataset contains 31 large instances with 1,002 to 7,397 cities, generated from six graphs. The number of salesmen is between 3 and 60. The 5 instances related to the first graph were presented in~\cite{Dong2018}, and the remaining instances in~\cite{Dong2019}.
\end{itemize}

We set the cut-off time $T_{max}$ for each instance of VSR-LKH-3 and LKH-3 for the CTSP the same as He et al.~\cite{He2021} does. That is, $T_{max}$ is set to 1, 10, and 60 minutes for \textit{Dataset \uppercase\expandafter{\romannumeral1}}, \textit{\uppercase\expandafter{\romannumeral2}}, and \textit{\uppercase\expandafter{\romannumeral3}}, respectively, except $T_{max}=240$ minutes for large instances with more than 7,000 cities. The parameters related to reinforcement learning in VSR-LKH-3 for the CTSP are set to the same as those in VSR-LKH-3 for the TSPTW (see Section \ref{sec_TSPTW_setup} for details).

\subsubsection{Comparison on VSR-LKH-3 with CTSP Baselines} 
\

The comparison results on VSR-LKH-3 with LKH-3 and GMA are shown in Table \ref{table_CTSP}. We compare the best and average solutions in 10 runs obtained by these algorithms. Column $m$ is the number of the salesmen. Column \textit{BKS} is the best-known solution reported by~\cite{He2021GMA,Helsgaun2017}. The values in brackets beside some results are the average violation values of these solutions. A star in column \textit{Best} indicates that a new best solution of the CTSP has been found. We also summarize the comparative results between VSR-LKH-3 and each baseline algorithm for the CTSP in Table \ref{table_Compare-CTSP}. The comparative results are based on all the 65 compared instances.

From the results in Tables \ref{table_CTSP} and \ref{table_Compare-CTSP}, we observe that:

(1) The best (average) solutions of VSR-LKH-3-$\alpha$ are better than those of LKH-3-$\alpha$ in 23 (29) instances, and worse than those of LKH-3-$\alpha$ in 4 (11) instances. The best (average) solutions of VSR-LKH-3-P are better than those of LKH-3-P in 21 (31) instances, and worse than those of LKH-3-P in 7 (10) instances. The results demonstrate that when solving the CTSP, VSR-LKH-3 also significantly outperforms LKH-3 with either the $\alpha$-measure or the POPMUSIC method.

(2) Both VSR-LKH-3-$\alpha$ and VSR-LKH-3-P significantly outperforms the advanced GMA heuristic algorithm for the CTSP. In particular, the performance of either LKH-3-$\alpha$ or LKH-3-P is weaker than GMA. Our proposed reinforcement learning method can help LKH-3 outperform GMA, indicating the excellent performance of our proposed method.

(3) VSR-LKH-3 provides 12 new best solutions for the 65 tested CTSP instances. In particular, VSR-LKH-3 provides the new best solutions for all the 5 large instances with more than 7,000 cities, indicating that our proposed algorithm is efficient and effective for large CTSP instances.

\begin{table}[t]
\centering
\caption{Comparison of VSR-LKH-3 and VSR-LKH-3-NoV.\vspace{-1em}}
\label{table_Ablation_VSRLKH3}
\scalebox{0.67}{\begin{tabular}{lrrrrr} \bottomrule
Algorithm\textbackslash{}Benchmark & \multicolumn{2}{c}{\textit{Dataset C}}          & \multicolumn{1}{c}{} & \multicolumn{2}{c}{\hspace{-1em}\textit{Dataset \uppercase\expandafter{\romannumeral3}}}  \\ \cline{2-3} \cline{5-6} 
                                   & Best            & Average              &                      & \hspace{-1em}Best        & Average            \\ \hline
VSR-LKH-3-NoV-$\alpha$             & (0.08) 12135.36 & (3.32) 12136.43      &                      & \hspace{-1em}29705510.29 & (0.24) 29802623.93 \\
VSR-LKH-3-$\alpha$                 & 12135.4         & (0.40) 12136.65      &                      & \hspace{-1em}29683345.97 & (0.15) 29731473.37 \\ \hline
VSR-LKH-3-NoV-P                    & 12138.4         & (163277.04) 89453.64 &                      & \hspace{-1em}29685164.71 & (0.61) 29711676.82 \\
VSR-LKH-3-P                        & 12137.5         & (77353.64) 46503.25  &                      & \hspace{-1em}29679655.81 & (0.29) 29701510.81 \\ \toprule
\end{tabular}}
\vspace{-1em}
\end{table}

\subsection{Ablation Study for VSR-LKH-3}
In order to demonstrate that our designed reward function in VSR-LKH-3 is effective and reasonable, we compare VSR-LKH-3 with its variant VSR-LKH-3-NoV for solving the TSPTW and CTSP. VSR-LKH-3-NoV is a variant of VSR-LKH-3 with the reward function in VSR-LKH (Eq. \ref{eq_r1}), i.e., without considering the violation function.

We compare VSR-LKH-3 with VSR-LKH-3-NoV in solving the TSPTW instances of \textit{Dataset C} and the CTSP instances of \textit{Dataset \uppercase\expandafter{\romannumeral3}}. The results are shown in Table \ref{table_Ablation_VSRLKH3}. Due to the limited space, we provide the average values of the best and average solutions of the algorithms. From the results we can see that with either the $\alpha$-measure or the POPMUSIC method, VSR-LKH-3 significantly outperforms VSR-LKH-3-NoV. The average violation values of the solutions obtained by VSR-LKH-3 are smaller than VSR-LKH-3-NoV, indicating that the designed reward function can help the algorithm find feasible solutions more easily, so as to find better solutions.


\section{Conclusion}
\label{Sec-Con}
We combine reinforcement learning with the powerful heuristic local search methods, LKH and LKH-3, for solving the TSP and its variant problems. We first propose a variable strategy reinforced algorithm called VSR-LKH for the TSP, and then propose another algorithm called VSR-LKH-3 that combines the variable strategy reinforcement learning method with LKH-3 for the TSP variants. These approaches use a Q-value as the metric for selecting and sorting the candidate cities, and allow the program to learn to select appropriate edges to be added from the candidate sets through reinforcement learning.

The variable strategy reinforced methods in VSR-LKH and VSR-LKH-3 combine the advantages of three reinforcement learning algorithms, Q-learning, Sarsa, and Monte Carlo, so as to further improve the flexibility and robustness of the proposed algorithms. VSR-LKH-3 further benefits from the new reward function based on the violation function for evaluating the infeasibility of the solutions, which enables the algorithm to find high-quality and feasible solutions more efficiently. Extensive experiments on public benchmarks show that with either the $\alpha$-measure or the POPMUSIC method for determining the candidate sets, VSR-LKH (VSR-LKH-3) significantly outperforms the LKH (LKH-3) algorithm for solving the TSPs. Moreover, our proposed reinforcement learning method can help LKH-3 outperform the state-of-the-art CTSP heuristic, creating 12 new records among the 65 public CTSP instances.

VSR-LKH and VSR-LKH-3 essentially reinforce the $k$-opt process of LKH and LKH-3. Hence, other algorithms based on $k$-opt could also be strengthened by our methods. Furthermore, our work demonstrates the feasibility and effectiveness of incorporating reinforcement learning with heuristic algorithms in solving various combinatorial optimization problems.

\bibliographystyle{IEEEtran}

\bibliography{main}
\end{document}